\documentclass{article}

\usepackage[preprint]{corl_2022} %

\usepackage{amsmath}
\usepackage{amssymb}
\usepackage{mathtools}
\usepackage{amsthm}
\usepackage{algorithmic}
\usepackage{algorithm}
\usepackage{xspace}
\usepackage{mathtools}
\mathtoolsset{showonlyrefs=true}

\usepackage{caption}
\usepackage{subcaption}
\usepackage{wrapfig}
\usepackage{sidecap}
\usepackage{multirow}
\usepackage{makecell}
\usepackage{booktabs}

\usepackage[capitalize,noabbrev]{cleveref}

\theoremstyle{plain}

\theoremstyle{definition}

\theoremstyle{remark}

\newcommand{\reb}[1]{\textcolor{black}{#1}}
\newcommand{\algo}{\emph{POIR}\xspace}
\newcommand{\wm}{$\hat{P}$} 
\newcommand{\matr}[1]{\mathbf{#1}}
\newcommand{\Comment}[1]{\textit{#1}}

\newcommand{\argmin}{\operatorname*{argmin}}
\newcommand{\argmax}{\operatorname*{argmax}}

\newcommand*\samethanks[1][\value{\thefootnote}]{\footnotemark[#1]}

\title{Get Back Here: Robust Imitation by Return-to-Distribution Planning}

\author{
  Geoffrey Cideron\\
  Google Research, Brain Team
  \And
   Baruch Tabanpour \\
   Google Research, Brain Team
   \And
   Sebastian Curi \\
   ETH
   \And
   Sertan Girgin\\
   Google Research, Brain Team
   \And
   Leonard Hussenot \\
   Google Research, Brain Team
   \And
   Gabriel Dulac-Arnold \\
   Google Research, Brain Team
   \And
   Matthieu Geist \\
   Google Research, Brain Team
   \And
   Olivier Pietquin\thanks{equal contribution} \\
   Google Research, Brain Team
   \And
   Robert Dadashi\samethanks[1] \\
   Google Research, Brain Team
}

\begin{document}
\maketitle

\begin{abstract}
    \looseness=-1
    We consider the Imitation Learning (IL) setup where expert data are not collected on the actual deployment environment but on a  different version. To address the resulting distribution shift, we combine behavior cloning (BC) with a planner that is tasked to bring the agent back to states visited by the expert whenever the agent deviates from the demonstration distribution. The resulting algorithm, \algo, can be trained offline, and leverages online interactions to efficiently fine-tune its planner to improve performance over time. We test \algo on a variety of human-generated manipulation demonstrations in a realistic robotic manipulation simulator and show robustness of the learned policy to different initial state distributions and noisy dynamics.
\end{abstract}


\section{Introduction}
\label{sec:introduction}

\looseness=-1
Imitation Learning (IL) is a paradigm in sequential decision making where an agent uses offline expert trajectories to mimic the expert's behavior~\citep{pomerleau1991efficient}. While Reinforcement Learning (RL) requires an additional reward signal that can be hard to specify in practice, IL only requires expert trajectories that can be easier to collect. In part due to its simplicity, IL has been applied successfully in several real world tasks, from robotic manipulation \citep{zhang2018deep,zeng2020transporter,florence2022implicit} to autonomous driving \citep{pomerleau1989alvinn,bojarski2016end}.\\
A key challenge in deploying IL, however, is that the agent may encounter states in the final deployment environment that were not labeled by the expert offline \citep{ross2011reduction}. In applications such as healthcare \citep{thomas2019preventing,gottesman2018evaluating} and robotics \citep{bousmalis2018using,kalashnikov2018qt}, online experimentation can be risky (\textit{e.g.}, on human patients) or costly to label (\textit{e.g.}, off-policy robotic datasets can take months to collect). In this work, we are thus interested in methods that are both robust in states not seen during training and also able to adapt online without access to a queryable expert.\\
Typical Imitation Learning approaches such as Behavior Cloning (BC)~\citep{pomerleau1991efficient} and Adversarial Imitation Learning (AIL)~\citep{ho2016generative} are not designed to be both data efficient and robust to distribution mismatch between the expert data and the final deployment environment. BC treats IL as a supervised learning problem, maximizing the likelihood of taking an expert action under the state distribution of the expert \citep{pomerleau1991efficient}. While BC generates useful policies from offline expert data, it performs poorly in states not seen during training \citep{ross2011reduction}: not only is BC brittle to out-of-distribution (OOD) inputs, it does not have access to a queryable expert and thus cannot adapt to new states once deployed. In contrast, AIL methods use an Inverse Reinforcement Learning (IRL) approach, first learning a reward function under which the expert trajectory data is optimal, and then training a policy using this learned reward function \citep{abbeel2004apprenticeship,ho2016generative,reddy2019sqil}. Since AIL methods train a Reinforcement Learning policy directly in the deployed environment, they can learn a policy in states not initially seen in the expert data. Nevertheless, AIL methods require large amounts of online data to reach expert performance and are typically hard to tune in practice \citep{kostrikov2019discriminator}, while BC uses no online data and is simpler to tune.\\
This paper proposes to take advantage of BC's data-efficiency and ability to train offline, while leveraging the flexibility of offline model-based methods~\citep{argenson2020model, kidambi2020morel, yu2020mopo} to be robust to out-of-distribution states and to fine-tune the final policy once deployed. \algo  (Planning from Offline Imitation Rewards), a Model-Based Imitation Learning algorithm, can learn a robust policy using an offline dataset of expert demonstrations (i.e. without any rewards), and continue to improve from data collected in the environment once deployed.  Through the use of an explicit imitation reward used during planning, \algo can return to states that were seen in the expert data if it ends up is states that are outside of the expert's support.\\
\algo can be deployed in both the offline and online settings. In the offline setting, the BC policy and the world model used by the planner are trained only on offline expert transitions. In the online setting, the world model is fine-tuned on new transitions collected by the planner policy in the deployed environment, improving the world model's prediction accuracy on parts of the state space not covered by offline transitions; this in turn improves the planner policy performance in out-of-distribution states with respect to the offline expert data.\\
Our contributions in this work are twofold: First, we introduce \algo, a Model-Based Imitation Learning method that is robust to both initial state distribution perturbations and stochasticity in the deployed environment. We demonstrate its performance compared to BC and AIL methods on a series of complex human-generated robotic manipulation demonstrations on the realistic robosuite~\citep{robosuite2020} simulator. Second, we show that \algo can be trained offline and deployed similarly to BC, while also continuing to improve its performance once online with significantly less data compared to AIL methods, avoiding the complexities of an RL training algorithm and adversarial loss.

\vspace{-0.25cm}
\section{Method}
\label{sec:method}
\vspace{-0.25cm}

We model environments as episodic Markov Decision Processes (MDPs) $(S, A, P, r, \rho_0, T)$  \citep{sutton2018reinforcement}, where $S$ denotes the state space, $A$ the action space, $P$ the transition kernel, $r$ the reward function, $\rho_0$ the initial state distribution, and $T$ the task horizon. A policy $\pi$ is a mapping from states to a distribution over actions; we denote the space of all policies as $\Pi$. Let $\rho_\pi$ denote the distribution over states and actions induced by policy $\pi$. In Imitation Learning, we do not know the true reward $r$. Instead we construct an imitation reward $r_{IL}$ offline based on the expert demonstrations $D_e=\{(s_t, a_t, s_{t+1})_{t\in T}\}$ induced by an expert policy $\pi_e$.

The goal of Imitation Learning is to find a policy $\pi$ such that its induced state-action distribution $\rho_\pi$ minimizes some divergence $D$ to the state-action distribution of the expert $\rho_{\pi_E}$ \citep{ghasemipour2020divergence}. In other words, the goal is to minimize: $\min_{\pi} D(\rho_\pi, \rho_{\pi_E})$. For example, the objective of GAIL~\citep{ho2016generative} is to minimize a regularized form of the Jensen-Shannon divergence. 

Next, we present \algo: a Model-based Imitation Learning algorithm that leverages an offline dataset of human expert demonstrations for robust imitation learning, while further improving with online data collected in the environment. \algo is composed of three main components: the BC policy prior, the planner, and the imitation reward.

\begin{wrapfigure}{o}{0.5\textwidth}
\vspace{-55pt}
\includegraphics[width=\linewidth]{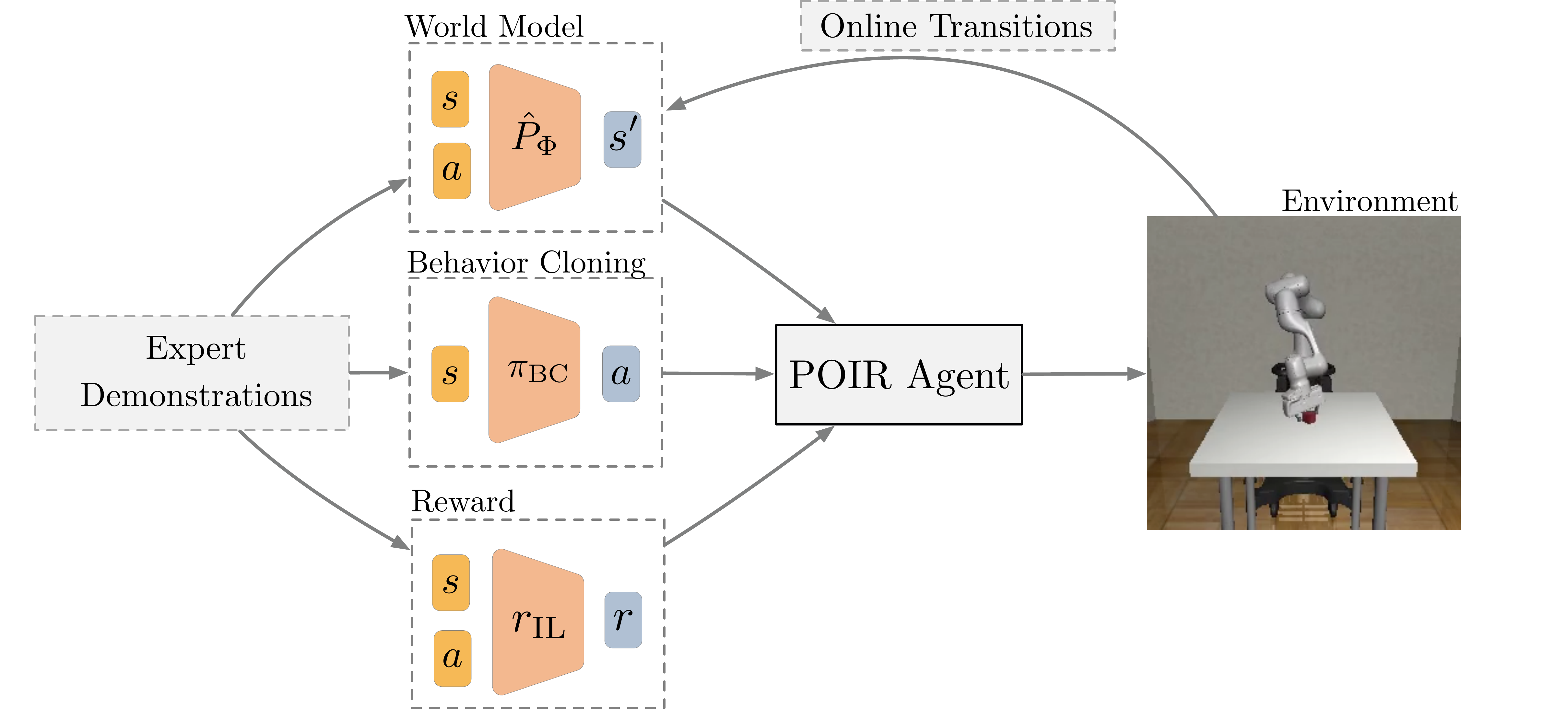}
\caption{Overview of various \algo components and their dependencies.  The BC policy and world model are trained from the expert data, whereas the imitation reward can depend either on the expert data, the world model's ensemble discrepancy, or the BC policy's ensemble discrepancy.}
\label{fig:algo-diagram}
\vspace{-10pt}
\end{wrapfigure}

\subsection{General Approach}
Our general approach extends the existing advantages inherent in BC with the ability of a planning algorithm to produce robust actions and adapt efficiently online. We do this by using the BC policy as a prior to guide trajectory rollouts in the planner. The planner uses a learnt world model to generate trajectories, and then selects the best action ranked according to a reward function; we leverage an imitation reward to incentivize the agent to stay close to the expert's support. All components of the system are initially trained on the dataset of expert trajectories $D_e=\{(s_t, a_t, s_{t+1})_{t\in T}\}$. Details about the three core components of \algo (the BC prior policy, the planner, and the imitation reward) are provided in the following sections and illustrated in Figure \ref{fig:algo-diagram}. An overview of \algo can be found in \cref{alg:mbip}.

\subsection{BC Policy Prior}
\label{sec:method-bc}
\vspace{-0.25cm}
\algo is a planning-based method that leverages BC as a prior to guide planner rollouts.  One key aspect of our work is that both the BC policy prior and the world model (Sec. \ref{sec:wm}) are implemented as bootstrap ensembles~\citep{lakshminarayanan2017simple,nagabandi2020deep,chua2018deep}. 
Our BC policy prior is in effect an ensemble of $K$ networks, which is used in a round-robin fashion in the planner as described in Sec. \ref{sec:planner}.
When using BC as a baseline, however, we have chosen to average the output of the $K$ networks to produce the final output action, which we refer to as Ensemble BC (EBC). This gives our Ensemble BC policy the following form: $\pi_{BC} = \frac{1}{K}\sum_1^K \pi_{\theta_k}(s)$, which shows significant performance gains over single-network BC (see Figure~\ref{on-off-init-noise-fig}), also observed in prior work~\citep{sasaki2020behavioral}. We train EBC on the full set of expert demonstrations for a given task, minimizing the action prediction error: $\theta_k = \argmin_{\theta_k}  \mathbb{E}_{D_e} [(\pi_{\theta_k}(s_t) - a_t)^2].$

\vspace{-0.5cm}
\subsection{Planner}
\label{sec:planner}
\vspace{-0.25cm}
The core of our algorithmic approach is to leverage a planner to return to the expert's state distribution. \algo uses Model-Predictive Control (MPC)~\citep{rault1978model} with an MPPI-based~\citep{williams2015model} trajectory optimizer, modified to support ensembles and policy priors as detailed by \cite{argenson2020model}. Intuitively, when the policy is already in-distribution relative to the expert demonstrations, the planner does not provide much added value. However when the policy is either strongly out of distribution, or decides to take actions that would result in leaving the distribution, the planner will provide candidate action trajectories that are closer to the demonstration distribution.  We provide more details on the learnt world model and the trajectory optimizer below.

\textbf{World Model}
\label{sec:wm}
A core component of planning-based approaches is the underlying world model. We use a deterministic function $\hat{P}_\phi : S \times A \rightarrow S$ parameterized by $\phi$, that takes a state-action pair and outputs the next state. It is learnt on the expert demonstration dataset $D_e$:
$\argmin_{\phi} \mathbb{E}_{(s_{t}, a_{t}, s_{t+1}) \in D_e}[(\hat{P}_\phi(s_t, a_t) - s_{t+1})^2]$. As mentioned previously, the world model is a bootstrap ensemble; $K$ world models are learned, where each $\hat{P}_k$ is trained on the same dataset but with different weight initializations. In the case of deterministic environments, the choice of a deterministic model vs. a stochastic one has been shown not to impact performance much~\citep{lutter2021learning}.  
Although certain experimental environments are stochastic, the demonstration data however is deterministic, and deterministic models have shown to work well in practice for fine-tuning.

\textbf{Planner trajectories}
\label{sec:planner_traj}
As described above, our policy leverages Model-Predictive Control (MPC)~\cite{rault1978model}, a simple yet effective control strategy~\citep{tassa2012synthesis, nagabandi2020deep}.  At each policy step, MPC queries a trajectory optimizer to find an optimal trajectory $A_H = \{a_1, \dots, a_H\}$ for the current state $s_t$, which maximizes some objective function $r$ according to the dynamics of a world model \wm.  MPC then applies the first action from the trajectory, $a_1$, to the system, and observes the corresponding next state $s_{t+1}$, with which the algorithm repeats.  
\algo leverages a modified version of MPPI~\citep{williams2017information, nagabandi2020deep, argenson2020model} which we now show in detail. \\
\algo's trajectory optimization scheme involves generating $N$ parallel trajectories sampled from a proposal distribution generated by $\pi_{BC}$. 
For each trajectory, an action is sampled and pertubed with gaussian noise, $\sim{a}_t = \{\pi_{BC}(s_t) + \epsilon_i\}_{i=1}^{N}$ with $\epsilon_i \sim \mathcal{N}(0, \sigma^2)$. 
The perturbed action $\tilde{a}_t$ is then passed into the world model to create the following $\hat{s}_{t+1} = \hat{P}(\hat{s}_{t}, \tilde{a}_{t})$. 
This procedure is repeated $H$ times to create a full action trajectory \reb{$A^n_H = \{\tilde{a}_1, \dots, \tilde{a}_H\}$}, in parallel for the $N$ trajectories.  The corresponding cumulative reward is also calculated to produce $R^n = \sum_{h=1}^H r_{IL}(\hat{s}^n_t, \reb{\tilde{a}^n_t})$ for each of the $N$ trajectories.\\
The final action returned, as per MPC, \reb{is the first action from the trajectory $i_{max}=\argmax_nR^n$ with the highest reward: $\pi_{\algo}(s_t) = A_{i_{max}}[0]$.} The full algorithm is detailed in Algorithm \ref{alg:mbip-trajopt}. 

\textbf{Use of Ensembles} As mentioned previously, both the policy prior $\pi_{BC}$ and the world model $\hat{P}$ are actually composed of $K$ ensembles. These are exploited by allocating each of the $N$ trajectories to one of the ensemble heads, both for the policy prior as well as the world model.  The allocation is constant throughout the rollout, as performed by ~\cite{nagabandi2020deep} and ~\cite{argenson2020model}, and allows for greater diversity both in action selection and model predictions.  For notational simplicity we don't describe this explicitly in Algorithm \ref{alg:mbip-trajopt}, but the full algorithm can be found in the Appendix, Algorithm \ref{alg:full_trajopt}.

\subsection{Imitation Reward}
\label{sec:imitation-reward}
The last component of \algo  is the imitation reward, $r_{IL}$. We construct several reward functions in the following sections, with the criteria that the reward should be high for states that are close to the expert support and low otherwise. The notion of proximity varies for each of our proposed rewards.

\textbf{L2 Reward} One simple way to construct such a reward function is to compute the L2 distance between the current state and all the states in the demonstration data, and then to take the minimum of these distances. This reward function does not require any pre-training. For a state $s$, we denote the L2 reward as: $r_{L2}(s) = - \min_{s_{e} \in D_{e}} ||s-s_{e}||_2$.

For large datasets, the L2 reward can become computationally expensive as we compute the distance to all states in the expert data. Approximate lookup approaches can reduce the complexity from $O(n)$ to $O(log(n))$ \citep{indyk1998approximate, guo2020accelerating}. The L2 reward, albeit simple, has shown to be effective on robotic arm manipulation tasks in previous works \citep{dadashi2021primal} and can be extended to higher dimensional state spaces by learning an embedding \cite{dadashi2021primal}. The variant of \algo using this reward function is called \algo-L2.  

\textbf{Ensemble-Disagreement Rewards} Another formulation of $r_{IL}$ is to consider the disagreement of the predictions over an ensemble of networks trained on the demonstration data.  Two well-known approaches use the disagreement of BC policies, as introduced in DRIL \cite{brantley2019disagreement}, and the disagreement of the world models, as introduced in MoREL~\citep{kidambi2020morel}. The former version is called \algo-DRIL while the latter is called \algo-MOREL. These approaches, when compared to \algo-L2, scale well with the size of the expert dataset and the state dimension, since they need only one forward pass during inference rather than an exhaustive search over $D_e$.  The rewards can be expressed as $r_{DRIL}(s) = - \max_{i,j} ||\pi_{\theta_i}(s)-\pi_{\theta_j}(s)||_{2}$ and $r_{MOREL}(s, a) = - \max_{i,j} ||\hat{P}_{\phi_i}(s, a)-\hat{P}_{\phi_j}(s, a)||_{2}$.

\vspace{-15pt}
\begin{minipage}{.55\linewidth}
\begin{algorithm}[H]
	 \caption{\algo}
     \label{alg:mbip}
    \begin{algorithmic}[1]
    \small
    \STATE {\bfseries Input:} $D_{e}$: expert trajectories, $L_{steps}$: \# of env. steps, $H$: planning horizon, $N$: number of trajectories, $\sigma$: policy prior noise scale.
    \STATE $\pi_{BC}$, $\hat{P}$, $r_{IL}$ = Train($D_{e}$)%
    \STATE $D_{a} = \{\}$
    \FOR{$t=1$ {\bfseries to} $L_{steps}$}
        \STATE Observe $s_t$
        \STATE $a_t$ = SelectAction$(s_t, \pi_{BC}, \hat{P}, r_{IL}, H, N, \sigma, k)$
        \STATE $s_{t+1}$ = EnvironmentStep($a_t$)
            \STATE $D_{a} = D_{a} \cup \{(s_t, a_t, s_{t+1})\}$
            \STATE Train $\hat{P}$ on $D_{e} \cup D_{a}$
    \ENDFOR
    \end{algorithmic}
\end{algorithm}
\end{minipage}
\begin{minipage}{.45\linewidth}
\begin{algorithm}[H]
	 \caption{\algo-SelectAction}
     \label{alg:mbip-trajopt}
    \begin{algorithmic}[1]
     \small
    \STATE {\bfseries SelectAction}($s, \pi_{BC}, \hat{P}, r_{IL}, H, N, \sigma$):
	    \STATE Set $\matr{R} = \vec{0}_N$ , $\matr{A} = \vec{0}_{N,H}$ 
	    \FOR {$n=1..N$} 
    	    \STATE $s_1 = s$
	        \FOR {$t=1..H$}
                \STATE $\epsilon \sim \mathcal{N}(0, \sigma^{2})$
                \STATE $\tilde{a}_t = \pi_{BC}(s_t) + \epsilon$ 
                \STATE $s_{t+1} = \hat{P}(s_t, \tilde{a}_t)$ 
                \STATE $\matr{A}_{n,t} = \tilde{a}_t$
                \STATE $R_n = R_n + r_{IL}((s_t, \tilde{a}_t, s_{t+1}))$
            \ENDFOR
        \ENDFOR
	    \RETURN $a_1 = A_{\argmax R, 1}$ %

    \end{algorithmic}
    \end{algorithm}
\end{minipage}
\vspace{-20pt}

\reb{\subsection{Online Fine-Tuning} We can perform online fine-tuning of \algo once deployed in an environment.  The main goal is to fine-tune \wm\  with additional data gained from interactions with the system.  
We can do this by simply augmenting the original dataset $D_e$ with data from the agent $D_a$ and training \wm on $D = D_e \cup D_a$.  
In the case where imitation rewards use ensemble discrepancy on \wm, we maintain a frozen version of \wm\ 
trained only on $D_e$ for the calculation of the imitation reward.}

\section{Experiments}
\label{sec:experiments}

To investigate the benefits of \algo in distribution mismatch settings, we train \algo on a series of human demonstration demonstrations on a realistic robotic manipulation tasks, and evaluate \algo's performance on various physically perturbed versions of the original environment. 
\vspace{-0.25cm}
\subsection{Environments and Demonstrations}
\textbf{Environments} We investigate \algo's performance on a series of robotic manipulation tasks defined in the Robosuite simulation environment \cite{robosuite2020}.  
We use these environments since they combine several key properties: the tasks are complex, a large amount of human generated demonstrations is available, and the environment authors have shown strong correlation between simulator performance and on-robot performance~\citep{mandlekar2021matters}. We focus on three tasks of increasing difficulty using the 7-DoF Panda arm model: \textbf{Lift}: grasp \& lift a cube off of the table (10-dim observation space), \textbf{PickPlaceCan}: grasp a can and place it in a bin (14-dim observation space), \textbf{NutAssemblySquare}: grasp a square nut and place it on a rod fixture (14-dim observation space).
All three tasks are deterministic and contain two sources of variation by default: the initial position of the robotic arm and the initial position of the object to grasp.

\begin{figure}[t]
\centering
\hfill
 \centering
 \includegraphics[width=\textwidth]{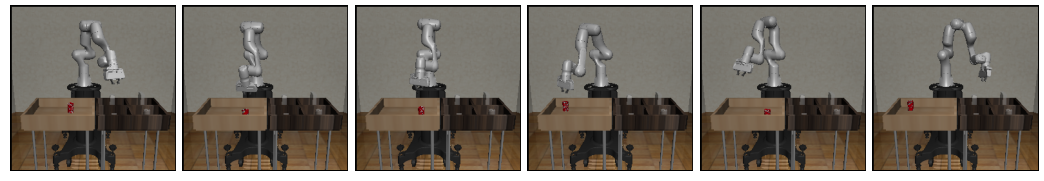}
\caption{
For noise of $\sigma_{noise}=0.4$ shown above, the arm can start in positions much further from the center of the table compared to default noise of $\sigma_{noise}=0.2$. Examples for all environments are shown in Appendix \ref{sec:init-noise-appendix}.}
\label{fig:env_robosuite_init_noise}
\vspace{-15pt}
\end{figure}

\textbf{Demonstrations} 
We use open-source human-generated demonstrations from the Robomimic~\citep{mandlekar2021matters} dataset, a well-studied dataset of human-generated demonstrations, for which there is strong evidence that approaches working well in the robosuite simulator also perform well on a real-robot version of the same tasks. 
For each task, we use the \textit{proficient human} demonstrations, composed of 200 demonstrations collected from a single proficient human operator using RoboTurk \citep{mandlekar2018roboturk}.

\subsection{Distribution mismatch}
The demonstrations are collected in an unperturbed version of the environment, which we refer to as the offline environment. We evaluate the performance of \algo on deployed environments that differ in two different manners: the diversity of the initial state distribution and the stochasticity in the transition dynamics of the environment.

\textbf{Initial state distribution.} For each task (Lift, PickPlaceCan, and NutAssemblySquare), we modify the initial state distribution of the robotic arm in the deployed environment. By default, the initial position of the arm is randomized around a mean position with a Gaussian noise of standard deviation $\sigma_{init}=0.02$. The default noise keeps the arm \textit{close} to the mean position at the beginning of each episode as depicted in Figure \ref{fig:env_robosuite_init_noise_multi}. Offline human demonstrations are collected with this default noise.
In our experiments, we gradually increase $\sigma_{init}$ in the deployed environment from $0.02$ to $0.4$. For larger $\sigma_{init}$ values, the initial position of the robotic arm can now be at either side of the table rather than centered directly above it, as shown in Figure \ref{fig:env_robosuite_init_noise}. The overall effect is illustrated in Appendix \ref{sec:init-noise-appendix} and in attached videos. The initial state noise modification leads to a distribution mismatch between the default initial positions in the demonstrations and the ones in the deployed environment.

\textbf{Stochasticity.} We introduce stochasticity to the otherwise deterministic Robosuite environments. Environment stochasticity can emerge in the real-world from naturally occurring variations in the controller or deployed environment \textit{e.g.}, friction, lighting, object deformations. We use a white Gaussian noise with standard deviation $\sigma_{action}$ as a simple model of these external perturbations. The Gaussian noise is added to the action taken by the agent at every step. Therefore the environment becomes stochastic, since two agents that take the same action at the same state may wind up in different next states.

\subsection{Implementation details}
\vspace{-0.25cm}
All algorithms are evaluated on $20$ episodes every $50,000$ environment steps, for a total of $500,000$ steps. Results for each method are averaged over $5$ random seeds. We report the average environment success rate $\in [0, 1]$ denoting whether or not the task was successfully completed. For each algorithm, the same set of hyperparameters are used across all environments. A full set of hyperparameters is reported in Appendix \ref{sec:hypers-appendix}. \\
Since \algo uses an ensemble of BC policies in the planner, BC and EBC are thus natural baselines.
We use the same network architecture (a multi-layer perceptron) for BC, EBC, and the BC policy used in \algo. For \algo and EBC, we use $27$ ensemble networks. For implementation reasons, ensembles are a multiple of 3, and we did not observe performance gains beyond 27.
In the planner, we sample $N=4000$ trajectories. We use $\sigma=0.2$ for the standard deviation noise around the BC action proposal. We use $H=5$ as the horizon for the planner trajectories as it performed best during initial hyperparameter searches. \\
We also compare our algorithm to state of the art imitation learning algorithms: Discriminator Actor-Critic (DAC) \citep{kostrikov2019discriminator} tuned using the guidelines of \cite{orsini2021matters}, ValueDice \citep{kostrikov2019imitation}, and SQIL \citep{reddy2019sqil}. DAC, ValueDice, and SQIL are natural baselines for \algo since they leverages both expert demonstrations and online data collected in the environment to learn its final policy. However, DAC and SQIL additionally differ from \algo as they cannot be used fully offline.\\
We find that state and action normalization helps \algo, but does not provide a benefit to BC or EBC. For DAC, we found that state normalization worked best. We normalize the offline expert demonstrations such that states and actions have mean $0$ and standard deviation $1$. Online transitions are normalized with the offline mean and standard deviation parameters.

\subsection{Results: Effects of initial state distribution mismatch}
\label{sec:results}

\begin{figure*}[ht]
     \centering
     \begin{subfigure}[b]{.9\textwidth}
         \centering
         \includegraphics[width=\textwidth]{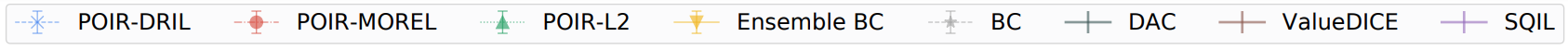}
     \end{subfigure}
     \hfill
     \begin{subfigure}[b]{.9\textwidth}
         \centering
         \includegraphics[width=\textwidth]{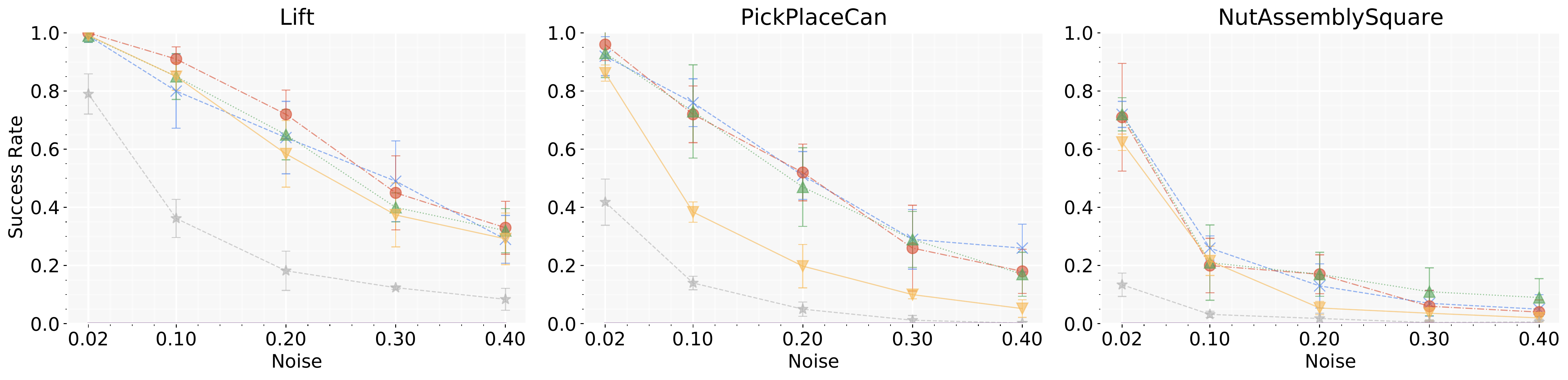}
         \caption{Offline performances}
         \label{offline-init-noise}
     \end{subfigure}
     \hfill
     \centering
     \begin{subfigure}[b]{.9\textwidth}
         \centering
         \includegraphics[width=\textwidth]{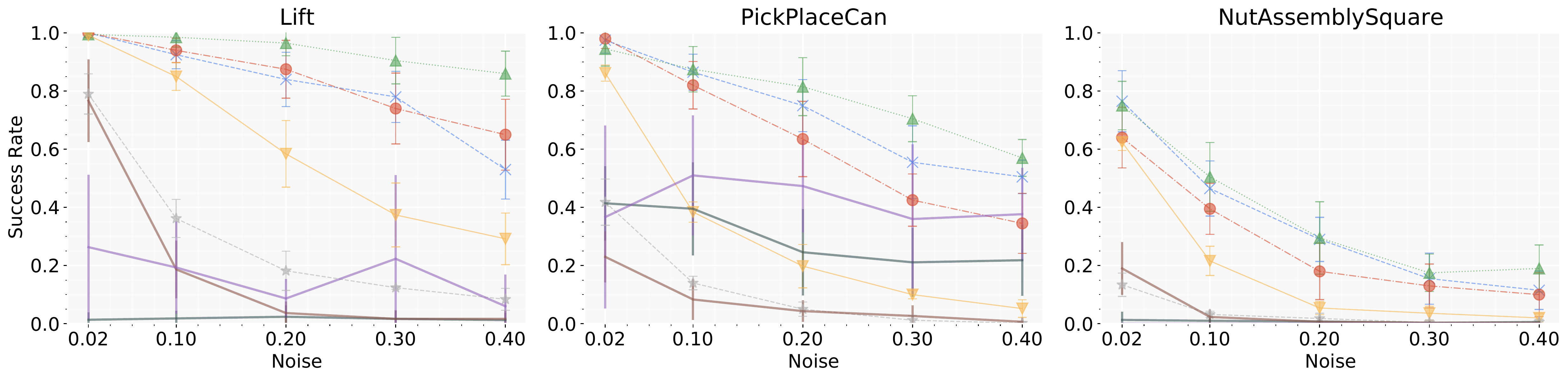}
         \caption{Online performances after 500,000 environment steps}
         \label{online-init-noise}
     \end{subfigure}
    \caption{Effects of $\sigma_{init}$ on task success rate for both off-line and on-line scenarios.}
    \label{on-off-init-noise-fig}
\end{figure*}

\textbf{Offline Performance.} In the fully offline setting, we observe that \algo have the best performances across the environments and noise level. Results for the offline setting are shown in the top row of the Figure~\ref{on-off-init-noise-fig}. For example in PickPlaceCan, Ensemble BC and \algo have nearly the same performance when there is no mismatch, but \algo has a $20\%-30\%$ increase in success rate for $\sigma_{init}=0.1$. This shows that the imitation rewards help the planner find better actions than the prior BC policy proposals in states not seen in the demonstrations.

\textbf{Online Performance.} \algo is able to efficiently leverage online data to improve its performance, as shown in the bottom row of Figure~\ref{on-off-init-noise-fig}. After online fine-tuning with $500,000$ environment steps, \algo-L2 performs best compared to other imitation rewards and BC/IL baselines. For example, there is a 50\% gap in success rate between \algo-L2 and EBC for Lift and PickPlaceCan for the largest value of noise ($\sigma_{init}=0.4$). This shows that improving the planner with data collected improves performance. This is a key difference between EBC and \algo as EBC cannot leverage this new data. RL-based approaches (DAC, ValueDICE, and SQIL) perform poorly as soon as some mismatch is introduced except for DAC on PickPlaceCan that achieves 20\% of success rate for $\sigma_{init}=0.4$ which is 3 times lower than \algo-L2 which achieve nearly 60\% of success rate in that configuration.

\textbf{Sample efficiency.} 
One of \algo's main advantages is that offline performance is competitive to BC and EBC baselines and that after an additional $500,000$ fine-tuning steps in the environment, \algo can also reach better performance compared to a state of the art IL method like DAC. In contrast, RL-based approaches (DAC, ValueDICE, and SQIL) are not able to reach BC performance within $500,000$ environment steps for Lift, and has a $0$ success rate even after $5M$ steps for NutAssemblySquare. This is illustrated in the Appendix: Figure~\ref{learning-curve-init-noise-fig2} shows the training curves for $5M$ environment steps.

\textbf{Emergent Retrying.} Qualitatively, we find that the BC agent moves towards the object of interest, attempts to grab it, and then continues to act as if the object were grasped  regardless of whether the grasp was actually successful. On the contrary, we notice in our experiments that \algo is able to retry grasping the object even after failing initial grasp attempts. We show emergent retrying for \algo in the attached video files as well as in the Appendix~\ref{sec:emergent-retry}. 

\reb{\textbf{Real time control.} Previous works showed that MPPI based approaches are amenable to real time control~\citep{williams2017information,nagabandi2020deep}. The overhead of POIR on top of MPPI consists in using a BC prior and computing a reward function which are computed with a forward pass on an ensemble of neural networks (and an additional lookup in the case of the L2 reward). We found experimentally that POIR-DRIL and POIR-MoREL run at 10 Hz and that POIR-L2 runs at 5Hz at inference on a TPUv2 GCP machine using a single core of the accelerator, which we argue shows the feasibility of the approach for real-time control.} 

\looseness=-1
\textbf{Ablations.} To showcase the importance of each component in \algo, we also ran two other baselines: \algo without a BC prior and PPO \citep{schulman2017proximal} with the DRIL reward which is essentially the DRIL algorithm \citep{brantley2019disagreement} without the auxiliary BC loss. \reb{The results are detailed in Appendix~\ref{sec:ablations}.} Both baselines yielded a 0\% success rate across all the experiments. These results show both the importance of the BC prior and the planner/IL reward combination.

\vspace{-0.25cm}
\subsection{Results: Effects of environment stochasticity mismatch}

\begin{figure*}[ht]
     \centering
     \begin{subfigure}[b]{.9\textwidth}
         \centering
         \includegraphics[width=\textwidth]{figures/legend.png}
     \end{subfigure}
     \hfill
     \begin{subfigure}[b]{.9\textwidth}
         \centering
         \includegraphics[width=\textwidth]{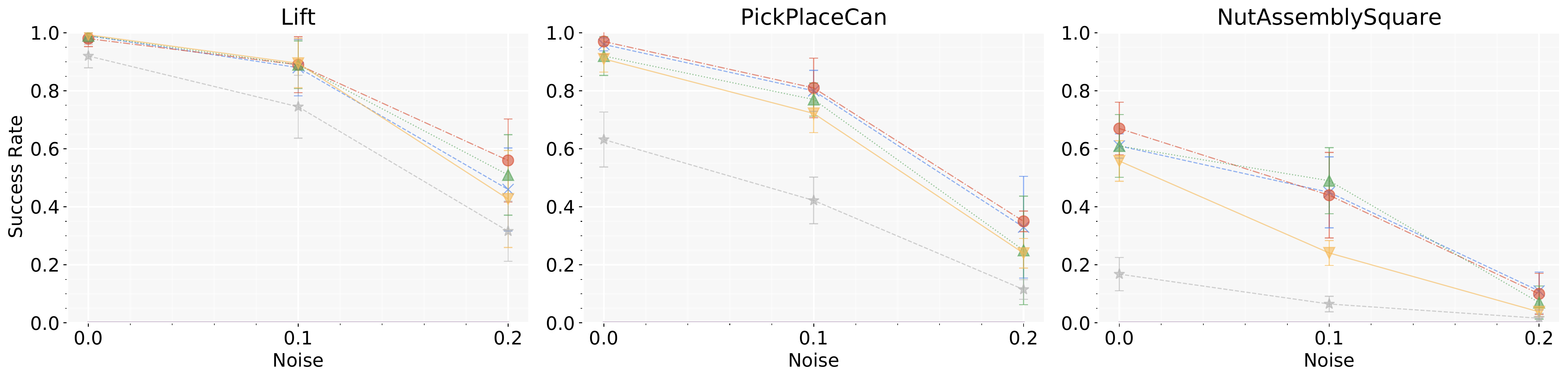}
         \caption{Offline performances}
         \label{fig:offline-action-noise}
     \end{subfigure}
     \hfill
     \centering
     \begin{subfigure}[b]{.9\textwidth}
         \centering
         \includegraphics[width=\textwidth]{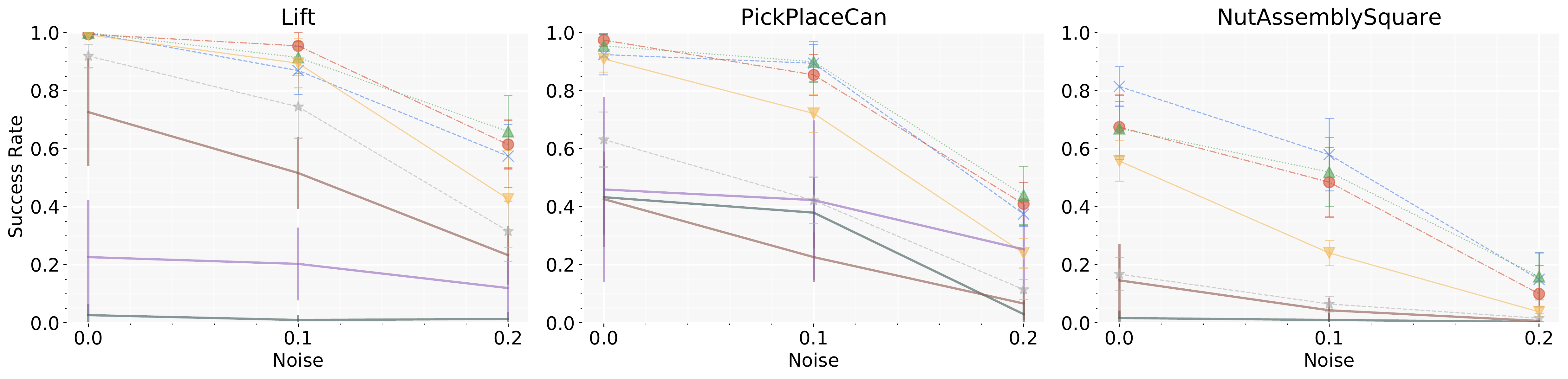}
         \caption{Online performances after 500,000 environment steps}
         \label{fig:online-action-noise}
     \end{subfigure}
    \caption{Effects of $\sigma_{action}$ on task success rate for both off-line and on-line scenarios. }
    \label{on-off-action-noise}
\end{figure*}

As shown in Figure \ref{on-off-action-noise}, \algo outperforms the other baselines when action noise stochasticity is added. We also remark that online data improves the trained policy. Notably with PickPlaceCan with an action noise of $0.1$, the gap in success rate between Ensemble BC and \algo-L2 is $25\%-35\%$. For NutAssemblySquare, the success rate gap between \algo-L2 and Ensemble BC goes from $0\%$ without action noise to nearly a $15\%$ success rate increase when the action noise is $0.1$. For Lift and NutAssemblySquare, DAC does not learn a good policy within the first $500,000$ environment steps. As shown in the Figure \ref{learning-curve-action-noise-fig2}, even with $5M$ environment steps, RL-based approaches perform worse than EBC in nearly all settings.

\vspace{-0.25cm}
\section{Related Work \& Limitations}
\vspace{-0.25cm}
Imitation Learning suffers from distribution shift between the expert data and the policy \citep{pomerleau1991efficient}. \cite{ross2010efficient} formalized this notion and showed that BC has a quadratic regret bound in $T$. Many IL methods have been designed to make policies more robust to distribution shift \citep{ross2011reduction,ng2000algorithms,abbeel2004apprenticeship}.

\textbf{IL with Queryable Expert Policy}: \cite{ross2011reduction} proposed DAGGER which queries the expert policy on online demonstrations to refine the BC policy and achieve linear regret in $T$. Recent studies use the policy uncertainty to selectively decide when to query the expert \citep{lee2018safe,silverio2019uncertainty,cui2019uncertainty,di2020safari}. Our IL method does not require a queryable expert or the explicit need for online demonstrations to mitigate distribution shift.

\textbf{Adversarial Imitation Learning}: AIL methods seek to match the state-action distribution of the policy to the fixed expert data by learning a discriminator that differentiates expert from non-expert transitions \citep{ho2016generative,kostrikov2019discriminator}. A model-free RL policy is trained on online environment trajectories by using the discriminator as a reward. MGAIL \citep{baram2017end} uses an additional world model to train a policy on exact gradients of the discriminator over entire trajectories. AIL methods generate a model-free controller that is hard to tune in practice and not sample efficient with respect to online data  \citep{ho2016generative}. We include comparisons to DAC, a state of the art AIL method, in Section \ref{sec:results}. In contrast to DAC, our method can be deployed offline and is more sample efficient online.

\textbf{IL with Support}: Recent IL methods explicitly use the distribution mismatch between the policy and expert as a reward signal for training a model-free controller, without the need for a discriminator \citep{dadashi2021primal,brantley2019disagreement,kim2018imitation,wang2019random,kelly2019hg}. These methods also require online data collection for training the RL policy, with similar sample efficiency as DAC.

\textbf{Offline RL}: Many offline RL methods use uncertainty estimates to make the final policy robust to distribution shift. Model-based offline RL methods leverage uncertainty in the world model or an ensemble of networks to keep the final policies close to the expert \citep{yu2020mopo,kidambi2020morel,yu2021combo,rafailov2021offline,argenson2020model,zhan2021model}. Offline RL differs from the IL setting as it assumes having access to the environment reward while in IL the environment reward is unknown. \reb{Offline RL methods are unadapted to deal with the IL rewards used by POIR. For instance, the L2 reward would be exactly 0 on the dataset of demonstrations.}

\textbf{Model-based Imitation Learning (MBIL)}: Using world models in Imitation Learning has been explored in other contexts \citep{osa2018algorithmic, englert2013probabilistic}. IMPLANT \citep{qi2022imitating} uses GAIL to learn a policy and value function using online data, which then get used in a planner policy. While IMPLANT shows that their planner is robust in noisy deployed environments for simple locomotion environments (Hopper, HalfCheetah, and Walker2d) our method differs mainly as we use BC instead of GAIL. This difference allows \algo to be deployed offline which is not the case of IMPLANT. Imitative Models \citep{rhinehart2018deep} learns a density model over expert data and then uses the density model as a reward in a goal conditioned open-loop planner; they demonstrate policies robust to noise in an autonomous driving simulator environment. A follow-up, Robust Imitative Planning (RIP) \citep{tigas2019robust}, uses an ensemble of networks to minimize uncertainty of trajectories in the planner, and then adapts the policy online by querying an expert policy \citep{filos2020can}. A similar method, RMBIL \citep{lin2021no} also shows improvements over BC when deployed in environments that mismatch the expert distribution. RMBIL is a purely offline method, that uses a neural ODE controller with a CVAE to encode states. In contrast to Imitative Models, RIP, RMBIL, and other similar works \citep{wu2020model}, our method uses a simpler policy: model-predictive control \citep{richalet1978model} with an open-loop trajectory planning algorithm \citep{argenson2020model} using BC priors; notably we show that \algo can run in both the offline and online settings without querying an expert. 

\textbf{Generalization in IL/RL}: Generalization in IL and RL seeks to improve performance on unseen interactions during deployment \citep{kirk2021survey}. Recent zero-shot and few-shot IL methods are designed to adapt to new tasks, objects, and domains using methods such as goal conditioning \citep{lynch2020learning,jang2021bc,james2018task}, meta-learning \citep{finn2017one,yu2018one}, domain randomization \citep{tobin2017domain,mehta2020active,xing2021kitchenshift}. Generalization improvements have also been shown in RL \citep{chebotar2021actionable,cobbe2019quantifying}. Our method explores the setting where the test environment is out-of-distribution compared to expert data, but not on new tasks or objects. We achieve improvements over BC on out-of-distribution test environments through a model-based controller, rather than with goal-conditioning or data augmentation.

\textbf{Limitations}: We foresee three limitations in our approach: As implemented in the paper, the L2 reward used in our experiments does not straightforwardly scale to large datasets, the method may struggle with higher-dimensional state spaces such as images, and experimental results were not validated on a physical robot.  In terms of scaling to large datasets, approaches exist to address this limitation such as approximate nearest-neighbour lookups~\citep{indyk1998approximate, guo2020accelerating}. For handling higher-dimensional observations such as pixels, approaches such as recurrent state-space models~\citep{hafner2019dream,hafner2019learning} or other embeddings~\citep{dadashi2021primal} where observations are embedded into a latent space could be used to create a more compact representation. Finally, although we did not test our approach on a real robot, evaluation of learning methods on the Robosuite simulator have been shown to correlate well with real-world robot experiments, in particular using the robomimic datasets~\citep{mandlekar2021matters} as we have done in this paper.
\vspace{-0.35cm}
\section{Conclusion}
\vspace{-0.35cm}
In summary, we present \algo, an Imitation Learning algorithm that combines BC with a planner. The BC policy creates candidate actions for the planner while the planner selects the best actions that will drive the agent back to states covered by the expert support. Unlike existing approaches, \algo uses a model-based controller that can be deployed either in the offline or online settings, without querying an expert. We demonstrate both the ability to learn entirely off-line, and downstream sample efficiency significantly better than common AIL methods.  We demonstrate that \algo generates actions that are significantly more robust to initial state distribution mismatch and stochastic actions in the deployed environment compared to AIL/BC. Finally, we empirically demonstrate that training the world model on online data further improves performance and  robustness of the policy to distribution mismatch.

\clearpage

\bibliography{example}  %

\begin{thebibliography}{76}
\providecommand{\natexlab}[1]{#1}
\providecommand{\url}[1]{\texttt{#1}}
\expandafter\ifx\csname urlstyle\endcsname\relax
  \providecommand{\doi}[1]{doi: #1}\else
  \providecommand{\doi}{doi: \begingroup \urlstyle{rm}\Url}\fi

\bibitem[Pomerleau(1991)]{pomerleau1991efficient}
D.~A. Pomerleau.
\newblock Efficient training of artificial neural networks for autonomous
  navigation.
\newblock \emph{Neural computation}, 3\penalty0 (1):\penalty0 88--97, 1991.

\bibitem[Zhang et~al.(2018)Zhang, McCarthy, Jow, Lee, Chen, Goldberg, and
  Abbeel]{zhang2018deep}
T.~Zhang, Z.~McCarthy, O.~Jow, D.~Lee, X.~Chen, K.~Goldberg, and P.~Abbeel.
\newblock Deep imitation learning for complex manipulation tasks from virtual
  reality teleoperation.
\newblock In \emph{2018 IEEE International Conference on Robotics and
  Automation (ICRA)}, pages 5628--5635. IEEE, 2018.

\bibitem[Zeng et~al.(2020)Zeng, Florence, Tompson, Welker, Chien, Attarian,
  Armstrong, Krasin, Duong, Sindhwani, et~al.]{zeng2020transporter}
A.~Zeng, P.~Florence, J.~Tompson, S.~Welker, J.~Chien, M.~Attarian,
  T.~Armstrong, I.~Krasin, D.~Duong, V.~Sindhwani, et~al.
\newblock Transporter networks: Rearranging the visual world for robotic
  manipulation.
\newblock \emph{arXiv preprint arXiv:2010.14406}, 2020.

\bibitem[Florence et~al.(2022)Florence, Lynch, Zeng, Ramirez, Wahid, Downs,
  Wong, Lee, Mordatch, and Tompson]{florence2022implicit}
P.~Florence, C.~Lynch, A.~Zeng, O.~A. Ramirez, A.~Wahid, L.~Downs, A.~Wong,
  J.~Lee, I.~Mordatch, and J.~Tompson.
\newblock Implicit behavioral cloning.
\newblock In \emph{Conference on Robot Learning}, pages 158--168. PMLR, 2022.

\bibitem[Pomerleau(1989)]{pomerleau1989alvinn}
D.~A. Pomerleau.
\newblock Alvinn: An autonomous land vehicle in a neural network.
\newblock Technical report, CMU, 1989.

\bibitem[Bojarski et~al.(2016)Bojarski, Del~Testa, Dworakowski, Firner, Flepp,
  Goyal, Jackel, Monfort, Muller, Zhang, et~al.]{bojarski2016end}
M.~Bojarski, D.~Del~Testa, D.~Dworakowski, B.~Firner, B.~Flepp, P.~Goyal, L.~D.
  Jackel, M.~Monfort, U.~Muller, J.~Zhang, et~al.
\newblock End to end learning for self-driving cars.
\newblock \emph{arXiv preprint arXiv:1604.07316}, 2016.

\bibitem[Ross et~al.(2011)Ross, Gordon, and Bagnell]{ross2011reduction}
S.~Ross, G.~Gordon, and D.~Bagnell.
\newblock A reduction of imitation learning and structured prediction to
  no-regret online learning.
\newblock In \emph{Proceedings of the fourteenth international conference on
  artificial intelligence and statistics}, pages 627--635. JMLR Workshop and
  Conference Proceedings, 2011.

\bibitem[Thomas et~al.(2019)Thomas, da~Silva, Barto, Giguere, Brun, and
  Brunskill]{thomas2019preventing}
P.~S. Thomas, B.~C. da~Silva, A.~G. Barto, S.~Giguere, Y.~Brun, and
  E.~Brunskill.
\newblock Preventing undesirable behavior of intelligent machines.
\newblock \emph{Science}, 366\penalty0 (6468):\penalty0 999--1004, 2019.

\bibitem[Gottesman et~al.(2018)Gottesman, Johansson, Meier, Dent, Lee,
  Srinivasan, Zhang, Ding, Wihl, Peng, et~al.]{gottesman2018evaluating}
O.~Gottesman, F.~Johansson, J.~Meier, J.~Dent, D.~Lee, S.~Srinivasan, L.~Zhang,
  Y.~Ding, D.~Wihl, X.~Peng, et~al.
\newblock Evaluating reinforcement learning algorithms in observational health
  settings.
\newblock \emph{arXiv preprint arXiv:1805.12298}, 2018.

\bibitem[Bousmalis et~al.(2018)Bousmalis, Irpan, Wohlhart, Bai, Kelcey,
  Kalakrishnan, Downs, Ibarz, Pastor, Konolige, et~al.]{bousmalis2018using}
K.~Bousmalis, A.~Irpan, P.~Wohlhart, Y.~Bai, M.~Kelcey, M.~Kalakrishnan,
  L.~Downs, J.~Ibarz, P.~Pastor, K.~Konolige, et~al.
\newblock Using simulation and domain adaptation to improve efficiency of deep
  robotic grasping.
\newblock In \emph{2018 IEEE international conference on robotics and
  automation (ICRA)}, pages 4243--4250. IEEE, 2018.

\bibitem[Kalashnikov et~al.(2018)Kalashnikov, Irpan, Pastor, Ibarz, Herzog,
  Jang, Quillen, Holly, Kalakrishnan, Vanhoucke, et~al.]{kalashnikov2018qt}
D.~Kalashnikov, A.~Irpan, P.~Pastor, J.~Ibarz, A.~Herzog, E.~Jang, D.~Quillen,
  E.~Holly, M.~Kalakrishnan, V.~Vanhoucke, et~al.
\newblock Qt-opt: Scalable deep reinforcement learning for vision-based robotic
  manipulation.
\newblock \emph{arXiv preprint arXiv:1806.10293}, 2018.

\bibitem[Ho and Ermon(2016)]{ho2016generative}
J.~Ho and S.~Ermon.
\newblock Generative adversarial imitation learning.
\newblock \emph{Advances in neural information processing systems},
  29:\penalty0 4565--4573, 2016.

\bibitem[Abbeel and Ng(2004)]{abbeel2004apprenticeship}
P.~Abbeel and A.~Y. Ng.
\newblock Apprenticeship learning via inverse reinforcement learning.
\newblock In \emph{Proceedings of the twenty-first international conference on
  Machine learning}, page~1, 2004.

\bibitem[Reddy et~al.(2019)Reddy, Dragan, and Levine]{reddy2019sqil}
S.~Reddy, A.~D. Dragan, and S.~Levine.
\newblock Sqil: Imitation learning via reinforcement learning with sparse
  rewards.
\newblock \emph{arXiv preprint arXiv:1905.11108}, 2019.

\bibitem[Kostrikov et~al.(2019)Kostrikov, Agrawal, Dwibedi, Levine, and
  Tompson]{kostrikov2019discriminator}
I.~Kostrikov, K.~K. Agrawal, D.~Dwibedi, S.~Levine, and J.~Tompson.
\newblock Discriminator-actor-critic: Addressing sample inefficiency and reward
  bias in adversarial imitation learning.
\newblock \emph{arXiv preprint arXiv:1809.02925}, 2019.

\bibitem[Argenson and Dulac-Arnold(2020)]{argenson2020model}
A.~Argenson and G.~Dulac-Arnold.
\newblock Model-based offline planning.
\newblock In \emph{International Conference on Learning Representations}, 2020.

\bibitem[Kidambi et~al.(2020)Kidambi, Rajeswaran, Netrapalli, and
  Joachims]{kidambi2020morel}
R.~Kidambi, A.~Rajeswaran, P.~Netrapalli, and T.~Joachims.
\newblock Morel: Model-based offline reinforcement learning.
\newblock In \emph{NeurIPS}, 2020.

\bibitem[Yu et~al.(2020)Yu, Thomas, Yu, Ermon, Zou, Levine, Finn, and
  Ma]{yu2020mopo}
T.~Yu, G.~Thomas, L.~Yu, S.~Ermon, J.~Y. Zou, S.~Levine, C.~Finn, and T.~Ma.
\newblock Mopo: Model-based offline policy optimization.
\newblock \emph{Advances in Neural Information Processing Systems},
  33:\penalty0 14129--14142, 2020.

\bibitem[Zhu et~al.(2020)Zhu, Wong, Mandlekar, and
  Mart\'{i}n-Mart\'{i}n]{robosuite2020}
Y.~Zhu, J.~Wong, A.~Mandlekar, and R.~Mart\'{i}n-Mart\'{i}n.
\newblock robosuite: A modular simulation framework and benchmark for robot
  learning.
\newblock In \emph{arXiv preprint arXiv:2009.12293}, 2020.

\bibitem[Sutton and Barto(2018)]{sutton2018reinforcement}
R.~S. Sutton and A.~G. Barto.
\newblock \emph{Reinforcement learning: An introduction}.
\newblock MIT press, 2018.

\bibitem[Ghasemipour et~al.(2020)Ghasemipour, Zemel, and
  Gu]{ghasemipour2020divergence}
S.~K.~S. Ghasemipour, R.~Zemel, and S.~Gu.
\newblock A divergence minimization perspective on imitation learning methods.
\newblock In \emph{Conference on Robot Learning}, pages 1259--1277. PMLR, 2020.

\bibitem[Lakshminarayanan et~al.(2017)Lakshminarayanan, Pritzel, and
  Blundell]{lakshminarayanan2017simple}
B.~Lakshminarayanan, A.~Pritzel, and C.~Blundell.
\newblock Simple and scalable predictive uncertainty estimation using deep
  ensembles.
\newblock In \emph{Advances in neural information processing systems}, pages
  6402--6413, 2017.

\bibitem[Nagabandi et~al.(2020)Nagabandi, Konolige, Levine, and
  Kumar]{nagabandi2020deep}
A.~Nagabandi, K.~Konolige, S.~Levine, and V.~Kumar.
\newblock Deep dynamics models for learning dexterous manipulation.
\newblock In \emph{Conference on Robot Learning}, pages 1101--1112, 2020.

\bibitem[Chua et~al.(2018)Chua, Calandra, McAllister, and Levine]{chua2018deep}
K.~Chua, R.~Calandra, R.~McAllister, and S.~Levine.
\newblock Deep reinforcement learning in a handful of trials using
  probabilistic dynamics models.
\newblock In \emph{Advances in Neural Information Processing Systems}, pages
  4754--4765, 2018.

\bibitem[Sasaki and Yamashina(2020)]{sasaki2020behavioral}
F.~Sasaki and R.~Yamashina.
\newblock Behavioral cloning from noisy demonstrations.
\newblock In \emph{International Conference on Learning Representations}, 2020.

\bibitem[Rault et~al.(1978)Rault, Richalet, Testud, and Papon]{rault1978model}
J.~Rault, A.~Richalet, J.~Testud, and J.~Papon.
\newblock Model predictive heuristic control: application to industrial
  processes.
\newblock \emph{Automatica}, 14\penalty0 (5):\penalty0 413--428, 1978.

\bibitem[Williams et~al.(2015)Williams, Aldrich, and
  Theodorou]{williams2015model}
G.~Williams, A.~Aldrich, and E.~Theodorou.
\newblock Model predictive path integral control using covariance variable
  importance sampling.
\newblock \emph{arXiv preprint arXiv:1509.01149}, 2015.

\bibitem[Lutter et~al.(2021)Lutter, Hasenclever, Byravan, Dulac-Arnold,
  Trochim, Heess, Merel, and Tassa]{lutter2021learning}
M.~Lutter, L.~Hasenclever, A.~Byravan, G.~Dulac-Arnold, P.~Trochim, N.~Heess,
  J.~Merel, and Y.~Tassa.
\newblock Learning dynamics models for model predictive agents.
\newblock \emph{arXiv preprint arXiv:2109.14311}, 2021.

\bibitem[Tassa et~al.(2012)Tassa, Erez, and Todorov]{tassa2012synthesis}
Y.~Tassa, T.~Erez, and E.~Todorov.
\newblock Synthesis and stabilization of complex behaviors through online
  trajectory optimization.
\newblock In \emph{2012 IEEE/RSJ International Conference on Intelligent Robots
  and Systems}, pages 4906--4913. IEEE, 2012.

\bibitem[Williams et~al.(2017)Williams, Wagener, Goldfain, Drews, Rehg, Boots,
  and Theodorou]{williams2017information}
G.~Williams, N.~Wagener, B.~Goldfain, P.~Drews, J.~M. Rehg, B.~Boots, and E.~A.
  Theodorou.
\newblock Information theoretic mpc for model-based reinforcement learning.
\newblock In \emph{2017 IEEE International Conference on Robotics and
  Automation (ICRA)}, pages 1714--1721. IEEE, 2017.

\bibitem[Indyk and Motwani(1998)]{indyk1998approximate}
P.~Indyk and R.~Motwani.
\newblock Approximate nearest neighbors: towards removing the curse of
  dimensionality.
\newblock In \emph{Proceedings of the thirtieth annual ACM symposium on Theory
  of computing}, pages 604--613, 1998.

\bibitem[Guo et~al.(2020)Guo, Sun, Lindgren, Geng, Simcha, Chern, and
  Kumar]{guo2020accelerating}
R.~Guo, P.~Sun, E.~Lindgren, Q.~Geng, D.~Simcha, F.~Chern, and S.~Kumar.
\newblock Accelerating large-scale inference with anisotropic vector
  quantization.
\newblock In \emph{International Conference on Machine Learning}, pages
  3887--3896. PMLR, 2020.

\bibitem[Dadashi et~al.(2021)Dadashi, Hussenot, Geist, and
  Pietquin]{dadashi2021primal}
R.~Dadashi, L.~Hussenot, M.~Geist, and O.~Pietquin.
\newblock Primal wasserstein imitation learning.
\newblock In \emph{ICLR 2021-Ninth International Conference on Learning
  Representations}, 2021.

\bibitem[Brantley et~al.(2019)Brantley, Sun, and
  Henaff]{brantley2019disagreement}
K.~Brantley, W.~Sun, and M.~Henaff.
\newblock Disagreement-regularized imitation learning.
\newblock In \emph{International Conference on Learning Representations}, 2019.

\bibitem[Mandlekar et~al.(2021)Mandlekar, Xu, Wong, Nasiriany, Wang, Kulkarni,
  Fei-Fei, Savarese, Zhu, and Mart{\'\i}n-Mart{\'\i}n]{mandlekar2021matters}
A.~Mandlekar, D.~Xu, J.~Wong, S.~Nasiriany, C.~Wang, R.~Kulkarni, L.~Fei-Fei,
  S.~Savarese, Y.~Zhu, and R.~Mart{\'\i}n-Mart{\'\i}n.
\newblock What matters in learning from offline human demonstrations for robot
  manipulation.
\newblock \emph{arXiv preprint arXiv:2108.03298}, 2021.

\bibitem[Mandlekar et~al.(2018)Mandlekar, Zhu, Garg, Booher, Spero, Tung, Gao,
  Emmons, Gupta, Orbay, et~al.]{mandlekar2018roboturk}
A.~Mandlekar, Y.~Zhu, A.~Garg, J.~Booher, M.~Spero, A.~Tung, J.~Gao, J.~Emmons,
  A.~Gupta, E.~Orbay, et~al.
\newblock Roboturk: A crowdsourcing platform for robotic skill learning through
  imitation.
\newblock In \emph{Conference on Robot Learning}, pages 879--893. PMLR, 2018.

\bibitem[Orsini et~al.(2021)Orsini, Raichuk, Hussenot, Vincent, Dadashi,
  Girgin, Geist, Bachem, Pietquin, and Andrychowicz]{orsini2021matters}
M.~Orsini, A.~Raichuk, L.~Hussenot, D.~Vincent, R.~Dadashi, S.~Girgin,
  M.~Geist, O.~Bachem, O.~Pietquin, and M.~Andrychowicz.
\newblock What matters for adversarial imitation learning?
\newblock \emph{arXiv preprint arXiv:2106.00672}, 2021.

\bibitem[Kostrikov et~al.(2019)Kostrikov, Nachum, and
  Tompson]{kostrikov2019imitation}
I.~Kostrikov, O.~Nachum, and J.~Tompson.
\newblock Imitation learning via off-policy distribution matching.
\newblock \emph{arXiv preprint arXiv:1912.05032}, 2019.

\bibitem[Schulman et~al.(2017)Schulman, Wolski, Dhariwal, Radford, and
  Klimov]{schulman2017proximal}
J.~Schulman, F.~Wolski, P.~Dhariwal, A.~Radford, and O.~Klimov.
\newblock Proximal policy optimization algorithms.
\newblock \emph{arXiv preprint arXiv:1707.06347}, 2017.

\bibitem[Ross and Bagnell(2010)]{ross2010efficient}
S.~Ross and D.~Bagnell.
\newblock Efficient reductions for imitation learning.
\newblock In \emph{Proceedings of the thirteenth international conference on
  artificial intelligence and statistics}, pages 661--668. JMLR Workshop and
  Conference Proceedings, 2010.

\bibitem[Ng et~al.(2000)Ng, Russell, et~al.]{ng2000algorithms}
A.~Y. Ng, S.~J. Russell, et~al.
\newblock Algorithms for inverse reinforcement learning.
\newblock In \emph{Icml}, volume~1, page~2, 2000.

\bibitem[Lee et~al.(2018)Lee, Saigol, and Theodorou]{lee2018safe}
K.~Lee, K.~Saigol, and E.~A. Theodorou.
\newblock Safe end-to-end imitation learning for model predictive control.
\newblock \emph{ArXiv}, abs/1803.10231, 2018.

\bibitem[Silv{\'e}rio et~al.(2019)Silv{\'e}rio, Huang, Abu-Dakka, Rozo, and
  Caldwell]{silverio2019uncertainty}
J.~Silv{\'e}rio, Y.~Huang, F.~J. Abu-Dakka, L.~Rozo, and D.~G. Caldwell.
\newblock Uncertainty-aware imitation learning using kernelized movement
  primitives.
\newblock In \emph{2019 IEEE/RSJ International Conference on Intelligent Robots
  and Systems (IROS)}, pages 90--97. IEEE, 2019.

\bibitem[Cui et~al.(2019)Cui, Isele, Niekum, and Fujimura]{cui2019uncertainty}
Y.~Cui, D.~Isele, S.~Niekum, and K.~Fujimura.
\newblock Uncertainty-aware data aggregation for deep imitation learning.
\newblock In \emph{2019 International Conference on Robotics and Automation
  (ICRA)}, pages 761--767. IEEE, 2019.

\bibitem[Di~Palo and Johns(2020)]{di2020safari}
N.~Di~Palo and E.~Johns.
\newblock Safari: Safe and active robot imitation learning with imagination.
\newblock \emph{arXiv preprint arXiv:2011.09586}, 2020.

\bibitem[Baram et~al.(2017)Baram, Anschel, Caspi, and Mannor]{baram2017end}
N.~Baram, O.~Anschel, I.~Caspi, and S.~Mannor.
\newblock End-to-end differentiable adversarial imitation learning.
\newblock In \emph{International Conference on Machine Learning}, pages
  390--399. PMLR, 2017.

\bibitem[Kim and Park(2018)]{kim2018imitation}
K.-E. Kim and H.~S. Park.
\newblock Imitation learning via kernel mean embedding.
\newblock In \emph{Thirty-Second AAAI Conference on Artificial Intelligence},
  2018.

\bibitem[Wang et~al.(2019)Wang, Ciliberto, Amadori, and
  Demiris]{wang2019random}
R.~Wang, C.~Ciliberto, P.~V. Amadori, and Y.~Demiris.
\newblock Random expert distillation: Imitation learning via expert policy
  support estimation.
\newblock In \emph{International Conference on Machine Learning}, pages
  6536--6544. PMLR, 2019.

\bibitem[Kelly et~al.(2019)Kelly, Sidrane, Driggs-Campbell, and
  Kochenderfer]{kelly2019hg}
M.~Kelly, C.~Sidrane, K.~Driggs-Campbell, and M.~J. Kochenderfer.
\newblock Hg-dagger: Interactive imitation learning with human experts.
\newblock In \emph{2019 International Conference on Robotics and Automation
  (ICRA)}, pages 8077--8083. IEEE, 2019.

\bibitem[Yu et~al.(2021)Yu, Kumar, Rafailov, Rajeswaran, Levine, and
  Finn]{yu2021combo}
T.~Yu, A.~Kumar, R.~Rafailov, A.~Rajeswaran, S.~Levine, and C.~Finn.
\newblock Combo: Conservative offline model-based policy optimization.
\newblock In \emph{Self-Supervision for Reinforcement Learning Workshop-ICLR
  2021}, 2021.

\bibitem[Rafailov et~al.(2021)Rafailov, Yu, Rajeswaran, and
  Finn]{rafailov2021offline}
R.~Rafailov, T.~Yu, A.~Rajeswaran, and C.~Finn.
\newblock Offline reinforcement learning from images with latent space models.
\newblock In \emph{Learning for Dynamics and Control}, pages 1154--1168. PMLR,
  2021.

\bibitem[Zhan et~al.(2021)Zhan, Zhu, and Xu]{zhan2021model}
X.~Zhan, X.~Zhu, and H.~Xu.
\newblock Model-based offline planning with trajectory pruning.
\newblock \emph{arXiv e-prints}, pages arXiv--2105, 2021.

\bibitem[Osa et~al.(2018)Osa, Pajarinen, Neumann, Bagnell, Abbeel, Peters,
  et~al.]{osa2018algorithmic}
T.~Osa, J.~Pajarinen, G.~Neumann, J.~A. Bagnell, P.~Abbeel, J.~Peters, et~al.
\newblock An algorithmic perspective on imitation learning.
\newblock \emph{Foundations and Trends in Robotics}, 7\penalty0 (1-2):\penalty0
  1--179, 2018.

\bibitem[Englert et~al.(2013)Englert, Paraschos, Deisenroth, and
  Peters]{englert2013probabilistic}
P.~Englert, A.~Paraschos, M.~P. Deisenroth, and J.~Peters.
\newblock Probabilistic model-based imitation learning.
\newblock \emph{Adaptive Behavior}, 21\penalty0 (5):\penalty0 388--403, 2013.

\bibitem[Qi et~al.(2022)Qi, Abbeel, and Grover]{qi2022imitating}
C.~Qi, P.~Abbeel, and A.~Grover.
\newblock Imitating, fast and slow: Robust learning from demonstrations via
  decision-time planning.
\newblock \emph{arXiv preprint arXiv:2204.03597}, 2022.

\bibitem[Rhinehart et~al.(2018)Rhinehart, McAllister, and
  Levine]{rhinehart2018deep}
N.~Rhinehart, R.~McAllister, and S.~Levine.
\newblock Deep imitative models for flexible inference, planning, and control.
\newblock \emph{arXiv preprint arXiv:1810.06544}, 2018.

\bibitem[Tigas et~al.(2019)Tigas, Filos, McAllister, Rhinehart, Levine, and
  Gal]{tigas2019robust}
P.~Tigas, A.~Filos, R.~McAllister, N.~Rhinehart, S.~Levine, and Y.~Gal.
\newblock Robust imitative planning: Planning from demonstrations under
  uncertainty.
\newblock In \emph{NeurIPS 2019 Workshop on Machine Learning for Autonomous
  Driving}, 2019.

\bibitem[Filos et~al.(2020)Filos, Tigkas, McAllister, Rhinehart, Levine, and
  Gal]{filos2020can}
A.~Filos, P.~Tigkas, R.~McAllister, N.~Rhinehart, S.~Levine, and Y.~Gal.
\newblock Can autonomous vehicles identify, recover from, and adapt to
  distribution shifts?
\newblock In \emph{International Conference on Machine Learning}, pages
  3145--3153. PMLR, 2020.

\bibitem[Lin et~al.(2021)Lin, Li, Zhou, Wang, and Meng]{lin2021no}
H.~Lin, B.~Li, X.~Zhou, J.~Wang, and M.~Q.-H. Meng.
\newblock No need for interactions: Robust model-based imitation learning using
  neural ode.
\newblock \emph{arXiv preprint arXiv:2104.01390}, 2021.

\bibitem[Wu et~al.(2020)Wu, Piergiovanni, and Ryoo]{wu2020model}
A.~Wu, A.~Piergiovanni, and M.~S. Ryoo.
\newblock Model-based behavioral cloning with future image similarity learning.
\newblock In \emph{Conference on Robot Learning}, pages 1062--1077. PMLR, 2020.

\bibitem[Richalet et~al.(1978)Richalet, Rault, Testud, and
  Papon]{richalet1978model}
J.~Richalet, A.~Rault, J.~Testud, and J.~Papon.
\newblock Model predictive heuristic control.
\newblock \emph{Automatica (journal of IFAC)}, 14\penalty0 (5):\penalty0
  413--428, 1978.

\bibitem[Kirk et~al.(2021)Kirk, Zhang, Grefenstette, and
  Rockt{\"a}schel]{kirk2021survey}
R.~Kirk, A.~Zhang, E.~Grefenstette, and T.~Rockt{\"a}schel.
\newblock A survey of generalisation in deep reinforcement learning.
\newblock \emph{arXiv preprint arXiv:2111.09794}, 2021.

\bibitem[Lynch et~al.(2020)Lynch, Khansari, Xiao, Kumar, Tompson, Levine, and
  Sermanet]{lynch2020learning}
C.~Lynch, M.~Khansari, T.~Xiao, V.~Kumar, J.~Tompson, S.~Levine, and
  P.~Sermanet.
\newblock Learning latent plans from play.
\newblock In \emph{Conference on Robot Learning}, pages 1113--1132. PMLR, 2020.

\bibitem[Jang et~al.(2021)Jang, Irpan, Khansari, Kappler, Ebert, Lynch, Levine,
  and Finn]{jang2021bc}
E.~Jang, A.~Irpan, M.~Khansari, D.~Kappler, F.~Ebert, C.~Lynch, S.~Levine, and
  C.~Finn.
\newblock Bc-z: Zero-shot task generalization with robotic imitation learning.
\newblock In \emph{5th Annual Conference on Robot Learning}, 2021.

\bibitem[James et~al.(2018)James, Bloesch, and Davison]{james2018task}
S.~James, M.~Bloesch, and A.~J. Davison.
\newblock Task-embedded control networks for few-shot imitation learning.
\newblock In \emph{Conference on Robot Learning}, pages 783--795. PMLR, 2018.

\bibitem[Finn et~al.(2017)Finn, Yu, Zhang, Abbeel, and Levine]{finn2017one}
C.~Finn, T.~Yu, T.~Zhang, P.~Abbeel, and S.~Levine.
\newblock One-shot visual imitation learning via meta-learning.
\newblock In \emph{Conference on Robot Learning}, pages 357--368. PMLR, 2017.

\bibitem[Yu et~al.(2018)Yu, Finn, Xie, Dasari, Zhang, Abbeel, and
  Levine]{yu2018one}
T.~Yu, C.~Finn, A.~Xie, S.~Dasari, T.~Zhang, P.~Abbeel, and S.~Levine.
\newblock One-shot imitation from observing humans via domain-adaptive
  meta-learning.
\newblock \emph{arXiv preprint arXiv:1802.01557}, 2018.

\bibitem[Tobin et~al.(2017)Tobin, Fong, Ray, Schneider, Zaremba, and
  Abbeel]{tobin2017domain}
J.~Tobin, R.~Fong, A.~Ray, J.~Schneider, W.~Zaremba, and P.~Abbeel.
\newblock Domain randomization for transferring deep neural networks from
  simulation to the real world.
\newblock In \emph{2017 IEEE/RSJ international conference on intelligent robots
  and systems (IROS)}, pages 23--30. IEEE, 2017.

\bibitem[Mehta et~al.(2020)Mehta, Diaz, Golemo, Pal, and
  Paull]{mehta2020active}
B.~Mehta, M.~Diaz, F.~Golemo, C.~J. Pal, and L.~Paull.
\newblock Active domain randomization.
\newblock In \emph{Conference on Robot Learning}, pages 1162--1176. PMLR, 2020.

\bibitem[Xing et~al.(2021)Xing, Gupta, Powers, and Dean]{xing2021kitchenshift}
E.~Xing, A.~Gupta, S.~Powers, and V.~Dean.
\newblock Kitchenshift: Evaluating zero-shot generalization of imitation-based
  policy learning under domain shifts.
\newblock In \emph{NeurIPS 2021 Workshop on Distribution Shifts: Connecting
  Methods and Applications}, 2021.

\bibitem[Chebotar et~al.(2021)Chebotar, Hausman, Lu, Xiao, Kalashnikov, Varley,
  Irpan, Eysenbach, Julian, Finn, et~al.]{chebotar2021actionable}
Y.~Chebotar, K.~Hausman, Y.~Lu, T.~Xiao, D.~Kalashnikov, J.~Varley, A.~Irpan,
  B.~Eysenbach, R.~Julian, C.~Finn, et~al.
\newblock Actionable models: Unsupervised offline reinforcement learning of
  robotic skills.
\newblock \emph{arXiv preprint arXiv:2104.07749}, 2021.

\bibitem[Cobbe et~al.(2019)Cobbe, Klimov, Hesse, Kim, and
  Schulman]{cobbe2019quantifying}
K.~Cobbe, O.~Klimov, C.~Hesse, T.~Kim, and J.~Schulman.
\newblock Quantifying generalization in reinforcement learning.
\newblock In \emph{International Conference on Machine Learning}, pages
  1282--1289. PMLR, 2019.

\bibitem[Hafner et~al.(2019{\natexlab{a}})Hafner, Lillicrap, Ba, and
  Norouzi]{hafner2019dream}
D.~Hafner, T.~Lillicrap, J.~Ba, and M.~Norouzi.
\newblock Dream to control: Learning behaviors by latent imagination.
\newblock \emph{arXiv preprint arXiv:1912.01603}, 2019{\natexlab{a}}.

\bibitem[Hafner et~al.(2019{\natexlab{b}})Hafner, Lillicrap, Fischer, Villegas,
  Ha, Lee, and Davidson]{hafner2019learning}
D.~Hafner, T.~Lillicrap, I.~Fischer, R.~Villegas, D.~Ha, H.~Lee, and
  J.~Davidson.
\newblock Learning latent dynamics for planning from pixels.
\newblock In \emph{International Conference on Machine Learning}, pages
  2555--2565, 2019{\natexlab{b}}.

\bibitem[Kingma and Ba(2014)]{kingma2014adam}
D.~P. Kingma and J.~Ba.
\newblock Adam: A method for stochastic optimization.
\newblock \emph{arXiv preprint arXiv:1412.6980}, 2014.

\bibitem[Hoffman et~al.(2020)Hoffman, Shahriari, Aslanides, Barth-Maron,
  Behbahani, Norman, Abdolmaleki, Cassirer, Yang, Baumli,
  et~al.]{hoffman2020acme}
M.~Hoffman, B.~Shahriari, J.~Aslanides, G.~Barth-Maron, F.~Behbahani,
  T.~Norman, A.~Abdolmaleki, A.~Cassirer, F.~Yang, K.~Baumli, et~al.
\newblock Acme: A research framework for distributed reinforcement learning.
\newblock \emph{arXiv preprint arXiv:2006.00979}, 2020.

\end{thebibliography}

\newpage
\appendix
\section{Learning Curves}
In this section, we show the learning curves until 5M of environment steps. 

\subsection{Initial State Distribution Mismatch.}

\begin{figure}[hbt!]
\begin{center}
\centerline{\includegraphics[width=\columnwidth]{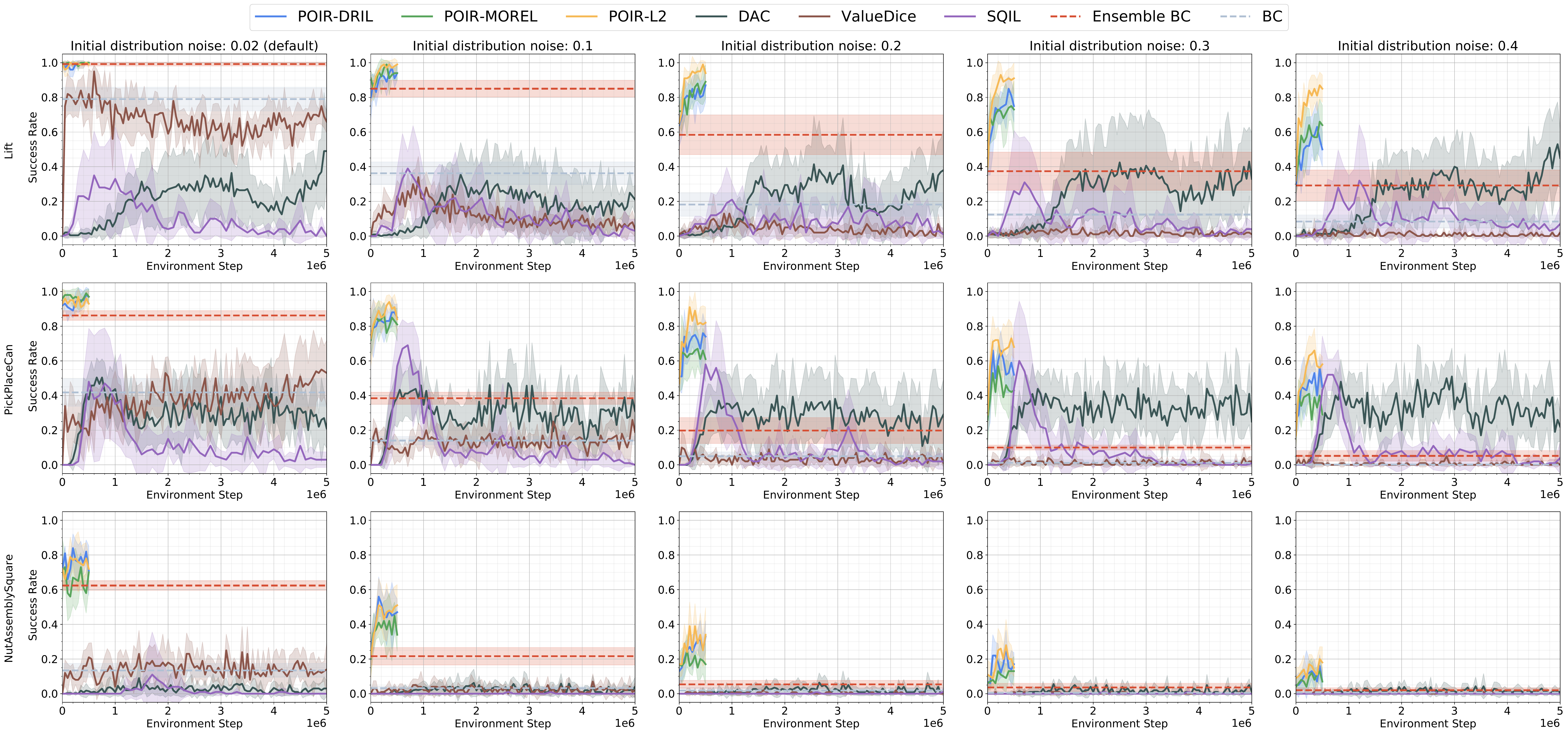}}
\caption{Effects of $\sigma_{init}$ on \algo success rate: \algo is able to maintain better performance compared to BC and DAC baselines across all environments and initial noise values. The performance gap between Ensemble BC and \algo widens as distribution mismatch increases. The shaded areas correspond to one standard deviation around the mean.}
\label{learning-curve-init-noise-fig2}
\end{center}
\end{figure}

\subsection{Action Noise.}

\begin{figure}[hbt!]
\begin{center}
\centerline{\includegraphics[width=\columnwidth-1cm]{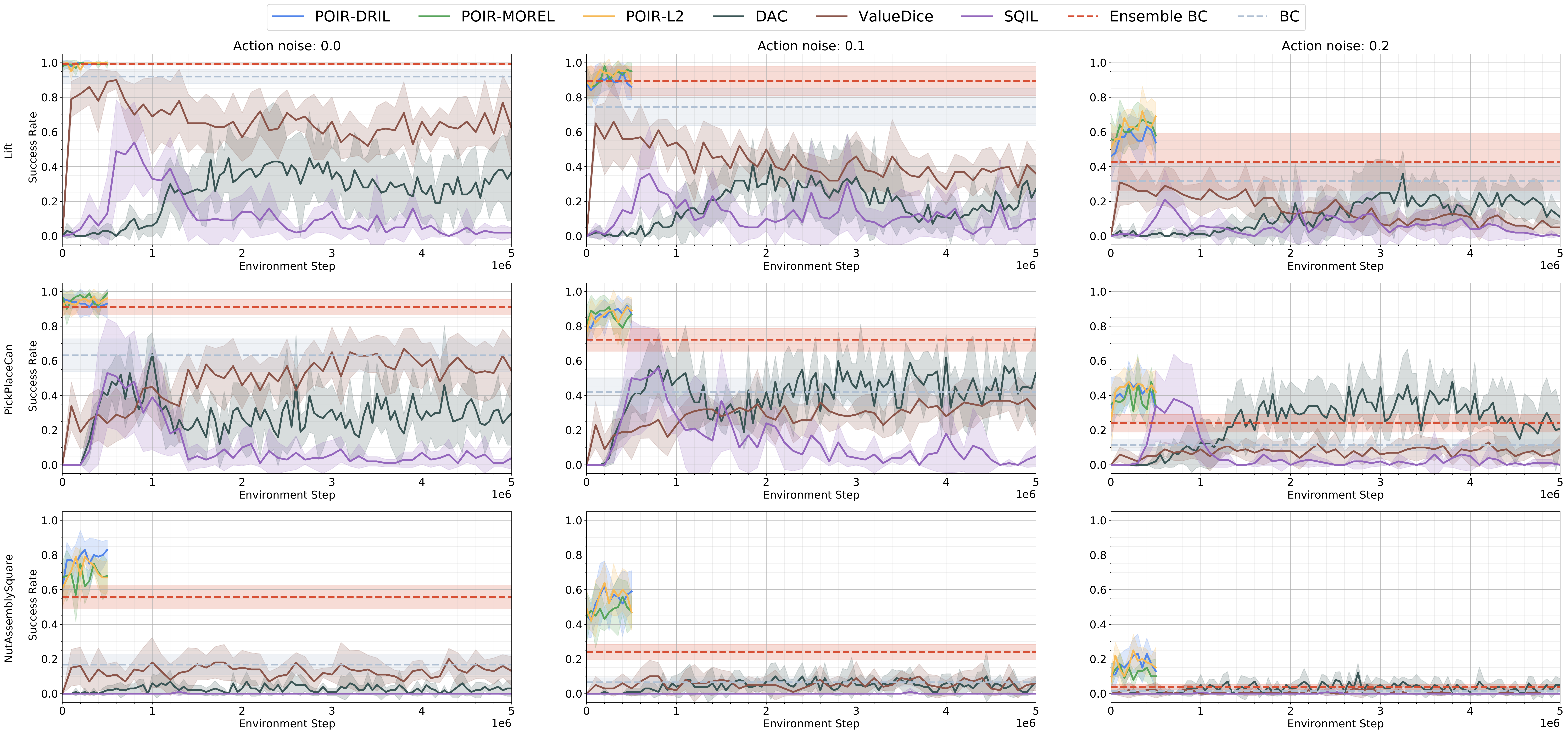}}
\caption{Effects of $\sigma_{action}$ on \algo success rate: This figure describes the effects of action noise on the various evaluated approaches.  We can observe that \algo methods are consistently better than other baselines. The shaded areas correspond to one standard deviation around the mean.}
\label{learning-curve-action-noise-fig2}
\end{center}
\end{figure}

\clearpage
\section{Ablations}
\label{sec:ablations}
\reb{We made several ablations in order to reflect the importance of each of the components of the method: the BC prior, the planner, and the IL reward. The ablation experiments of Table-\ref{ablations-lift}-\ref{ablations-pickplacecan}-\ref{ablations-nutassemblysquare} show that each of the components is crucial to the performance of POIR:
\begin{itemize}
    \item Ensemble BC: this algorithm can be seen as an ablation of our approach as it only relies on the BC prior. The results show that Ensemble BC performs strongly without any distribution mismatch but its performance quickly degrades as the mismatch increases. Therefore, on the expert support, the BC prior selects actions good enough to solve the task but it does not recover from unseen states. 
    \item POIR without a BC prior: in this version, the planner does not rely on BC to pick action candidates and instead samples actions uniformly at random. This leads to 0\% success rate across every environments hence showing that the BC prior is crucial.
    \item PPO + DRIL reward: in this version, we trained the PPO algorithm using the public implementation from acme\footnote{https://github.com/deepmind/acme/tree/master/acme/agents/jax/ppo} algorithm to maximize one of the proposed rewards (DRIL). This algorithm yields a 0\% success rate across the environments. This shows that the signal from the IL reward alone is not informative enough to solve the task even with an online RL algorithm. However, POIR's specific combination (BC prior + planner + IL Reward) leverages the reward signal to find the actions that will lead the agent to stay on the expert support where BC is a strong prior.
\end{itemize}
These ablations show that each component alone is not enough to tackle the distribution mismatch problem: BC cannot recover from unseen states; the planner can only leverage reasonable candidate actions with limited compute; the IL reward alone is too weak of a signal to lead to task completion.
}

\begin{table*}[h]
  \centering
  {\color{blue}\begin{tabular}{|c|c|c|c|c|c|}
    \hline
    Magnitude & 0.02 & 0.1 & 0.2 & 0.3 & 0.4 \\
    \hline
    \hline
    \makecell{POIR-DRIL \\(online)} & 1 & 0.9 & 0.78 & 0.65 & 0.52  \\
    \hline
    \makecell{POIR-DRIL \\(offline)}  & 1 & 0.77 & 0.65 & 0.4 & 0.22 \\
    \hline
    \makecell{Ensemble BC}  & 1 & 0.85 & 0.60 & 0.38 & 0.3  \\  
    \hline
    PPO-DRIL   & 0 & 0 & 0 & 0 & 0 \\
    \hline
    \makecell{POIR-DRIL \\ (without BC prior)}  & 0 & 0 & 0 & 0 & 0   \\ 
    \hline
  \end{tabular}}
  \caption{\reb{Ablation of the components of the POIR algorithm on Lift}}
  \label{ablations-lift}
\end{table*}

\begin{table*}[h]
  \centering
  {\color{blue}\begin{tabular}{|c|c|c|c|c|c|}
    \hline
    magnitude & 0.02 & 0.1 & 0.2 & 0.3 & 0.4 \\
    \hline
    \hline
    \makecell{POIR-DRIL \\(online)} & 0.92 & 0.83 & 0.65 & 0.5 & 0.4  \\
    \hline
    \makecell{POIR-DRIL \\(offline)}  & 0.96 & 0.75 & 0.5 & 0.33 & 0.25 \\
    \hline
    \makecell{Ensemble BC}  & 0.87 & 0.38 & 0.2 & 0.1 & 0.05 \\  
    \hline
    PPO-DRIL   & 0 & 0 & 0 & 0 & 0  \\ 
    \hline
    \makecell{POIR-DRIL \\ (without BC prior)}  & 0 & 0 & 0 & 0 & 0 \\ 
    \hline
  \end{tabular}}
  \caption{\reb{Ablation of the components of the POIR algorithm on PickPlaceCan.}}
  \label{ablations-pickplacecan}
\end{table*}

\begin{table*}[h]
  \centering
  {\color{blue}\begin{tabular}{|c|c|c|c|c|c|}
    \hline
    magnitude & 0.02 & 0.1 & 0.2 & 0.3 & 0.4 \\
    \hline
    \hline
    \makecell{POIR-DRIL \\(online)} & 0.72 & 0.38 & 0.22 & 0.15 & 0.08   \\
    \hline
    \makecell{POIR-DRIL \\(offline)}  & 0.72 & 0.33 & 0.17 & 0.08 & 0.03 \\
    \hline
    \makecell{Ensemble BC}  & 0.62 & 0.22 & 0.05 & 0.03 & 0.02      \\  
    \hline
    PPO-DRIL   & 0 & 0 & 0 & 0 & 0  \\ 
    \hline
    \makecell{POIR-DRIL \\ (without BC prior)}  & 0 & 0 & 0 & 0 & 0 \\ 
    \hline
  \end{tabular}}
  \caption{\reb{Ablation of the components of the POIR algorithm on NutAssemblySquare.}}
  \label{ablations-nutassemblysquare}
\end{table*}

\newpage
\section{Hyperparameters \& Implementation Details}
\label{sec:hypers-appendix}

In this section, we outline the full set of hyperparameters for \algo, BC, and DAC.

\subsection{Details for BC, EBC, \& \algo.}

\paragraph{Offline Training.} For offline training of the BC policy and world models, we use $500$ training epochs, with each epoch doing a full pass on the offline expert trajectories with a batch size of $256$. \algo-DRIL and \algo-MOREL train the imitation reward policy and world models respectively on $50000$ gradient steps on offline data with batch size of $256$. We use an Adam optimizer \citep{kingma2014adam} with a learning rate of 1e-4, $\beta_1=0.9$, and $\beta_2=0.999$ for all networks.

\paragraph{\algo Online Training.} For  \algo online training, we use a replay buffer to store a mixture of expert data and online planner policy data $D = D_e \cup D_a$. We denote the ratio of online to offline data sampled from the buffer as the $mixture\_ratio$. At the beginning of online fine-tuning, we use a $mixture\_ratio=0.0$, and we linearly increase the ratio to a maximum of $0.5$ by an increment of $0.05$ for every 100 gradient steps. We train online for up to $500k$ environment steps.

\paragraph{Planner} Previous works based on MPPI~\citep{williams2017information, nagabandi2020deep,argenson2020model} use a softmax-averaging of the action trajectories according to their respective return, but this can be unstable in the face of large returns. We observe that top-k averaging performs equivalently experimentally while being more numerically stable.

\paragraph{Hyperparameters.} The networks for BC and world models use an MLP with $5$ layers, $relu$ activation functions, and $300$ hidden units. We use a single network for BC and an ensemble of $27$ networks for EBC and world models. We swept over parameters as shown in Table \ref{tab:main-hypers}, with final values shown in the right-most column. For BC and EBC, we found that normalizing states and actions did not provide a benefit. For \algo, we use both state and action normalization.

\begin{table}[h]
\centering
\tiny
\begin{tabular}{@{}r|r|r@{}}
\toprule
Parameter	& Sweep & Final Value \\ \midrule
\hline
Learning rate & \{1e-4, 5e-4\}  & 1e-4  \\
Hidden units  & \{100, 300, 500\} & 300 \\
Layers        & \{1, 3, 5\} & 5  \\
Num Networks        & \{1, 3, 9, 27\}   & 27    \\
Activation    & $\{relu, tanh\}$ & $relu$   \\
Planner: Num trajectories $N$ & \{1000, 4000\} & 4000 \\
Planner: Action Noise $\sigma$ & \{0.1, 0.15, 0.2\} & 0.2 \\
Planner: Horizon $H$ & \{1, 5, 10, 20\} & 5 \\
\algo: Normalize observations & \{True, 
False\} & True \\
\algo: Normalize actions & \{True, 
False\} & True \\
BC: Normalize observations & \{True, 
False\} & False \\
BC: Normalize action & \{True, 
False\} & False \\
\hline
\end{tabular}
\vspace{0.5cm}
\caption{Hyperparameters for EBC, the world model in \algo, and the planner. BC uses the same network hyperparameters, but with a single network only. Both BC and EBC do not make use of the planner.}
\label{tab:main-hypers}
\end{table}

\reb{\subsection{Details for the baselines (DAC, ValueDICE, SQIL, PPO+DRIL).} 
\paragraph{Hyperparameter selection.} For each baseline and hyperparameter set, we ran the algorithm on Lift with 3 initialization noises $\sigma \in \{0.02, 0.2, 0.4\}$. The hyperparameters that performed best where then used for the evaluation on all environments and noise perturbation value.
\subsubsection{DAC.}
For DAC, we used as reference the hyperparemeter configuration from \citet{orsini2021matters} and performed tuning on top of it. A full set of hyperparameters is shown in Table \ref{tab:dac-hypers}. The details of the different hyperparameters can be found here: \url{https://github.com/deepmind/acme/tree/master/acme/agents/jax/ail}.}

\begin{table}[h]
\centering
\tiny
{\color{blue}\begin{tabular}{c|c|c}
\hline
Parameter	& Sweep & Final Value \\ %
\hline
policy MLP depth & 2 & 2 \\
policy MLP width & 256 & 256\\
critic MLP depth & 3 & 3\\
critic MLP width & 512 & 512\\
activation & $relu$ & $relu$\\
discount & 0.97 & 0.97\\
batch size & 256 & 256\\
RL Algorithm & SAC & SAC \\
SAC learning rate &  \{1e-4, 3-e4\} & 3e-4  \\
SAC entropy dimension & -0.5  & -0.5\\
n step return & \{1, 3\} & 3 \\
replay buffer size & 3e6  & 3e6\\
Subtract logpi & False & False\\
Absorbing state & True & True\\
Discriminator input & $(s, a)$ & $(s, a)$\\
Discriminator MLP depth & 2 & 2\\
Discriminator MLP width & 128 & 128\\
Discriminator activation & $relu$ & $relu$\\
Gradient penalty coefficient & \{1, 10\} & 10\\
Normalize observations & \{False, True\} & True \\
Learning rate & 3e-5 & 3e-5\\
Reward function & $ln(D)-ln(1-D)$ & $ln(D)-ln(1-D)$ \\ 
\hline
\end{tabular}}

\vspace{0.5cm}
\caption{\reb{Hyperparameters for DAC.}}
\label{tab:dac-hypers}
\end{table}

\reb{
\subsubsection{ValueDice}
We used the reference implementation from the Acme library \citep{hoffman2020acme}. The hyperparameter sweep and the selected hyperparameters can be found in Table~\ref{tab:value-dice-hp}. The details of the different hyperparameters can be found here: \url{https://github.com/deepmind/acme/tree/master/acme/agents/jax/value_dice}.
}

\begin{table}[h]
\centering
\tiny
{\color{blue}\begin{tabular}{c|c|c}
\hline
Parameter	& Sweep & Final Value \\ %
\hline
nu\_learning\_rate & \{0.001, 0.03, 0.05\} & 0.01 \\
nu\_learning\_rate & \{0, 3, 10\} & 10 \\
policy\_learning\_rate & \{1e-5, 3e-5, 1e-4\} & 1e-5 \\
policy\_reg\_scale & \{0, 1e-5, 1e-4, 1e-3\}& 0.0001 \\
$\alpha$ & \{0, 0.05\}& 0.05 \\
batch\_size & 256 & 256\\
policy MLP depth & 2 & 2 \\
policy MLP width & 256 & 256 \\
nu MLP depth & 2 & 2 \\
nu MLP width & 256 & 256 \\

\end{tabular}
}
\vspace{0.5cm}
\caption{\reb{Hyperparameters for ValueDICE.}}
\label{tab:value-dice-hp}
\end{table}

\reb{
\subsubsection{SQIL}
We used the reference implementation from the Acme library \citep{hoffman2020acme}. The hyperparameter sweep and the selected hyperparameters can be found in Table~\ref{tab:sqil-hp}. The details of the different hyperparameters can be found here: \url{https://github.com/deepmind/acme/tree/master/acme/agents/jax/sac}.
}

\begin{table}[h]
\centering
\tiny
{\color{blue}\begin{tabular}{c|c|c}
\hline
Parameter	& Sweep & Final Value \\ %
\hline
policy MLP depth & 2 & 2 \\
policy MLP width & 256 & 256\\
critic MLP depth & 3 & 3\\
critic MLP width & 512 & 512\\
activation & $relu$ & $relu$\\
discount & 0.97 & 0.97\\
batch size & 256 & 256\\
RL Algorithm & SAC & SAC \\
SAC learning rate &  \{1e-4, 3-e4, 1e-3\} & 1e-4  \\
SAC entropy dimension & -0.5  & -0.5\\
n step return & \{1, 3\} & 3 \\
replay buffer size & 1e6  & 1e6\\
\end{tabular}
}
\vspace{0.5cm}
\caption{\reb{Hyperparameters for SQIL.}}
\label{tab:sqil-hp}
\end{table}

\reb{
\subsubsection{PPO+DRIL}
We used the reference implementation from the Acme library \citep{hoffman2020acme}. The hyperparameter sweep and the selected hyperparameters can be found in Table~\ref{tab:ppo-dril-hp}. The details of the different hyperparameters for PPO can be found here: \url{https://github.com/deepmind/acme/tree/master/acme/agents/jax/ppo}. As every configurations yielded a success rate of 0, we could not discriminate the impact of the different hyperparameters. 
}
\begin{table}[h]
\centering
\tiny
{\color{blue}\begin{tabular}{c|c|c}
\hline
Parameter	& Sweep & Final Value \\ %
\hline
policy MLP depth & 5 & 5 \\
policy MLP width & 256 & 256\\
activation & $relu$ & $relu$\\
PPO learning rate &  \{1e-4, 3-e4, 1e-3\} & 1e-4  \\
PPO entropy cost & \{0, 3e-4\} \\
PPO num\_epochs & 2 & 2 \\
PPO num\_minibatches & 8 & 8 \\
PPO unroll\_length & 8 & 8 \\
PPO value\_loss\_coef & 1 & 1 \\
DRIL num networks & 27 & 27 \\
DRIL batch size & 256 & 256 \\ 
DRIL num epochs & 10 & 10 \\ 
\end{tabular}
}
\vspace{0.5cm}
\caption{\reb{Hyperparameters for PPO+DRIL.}}
\label{tab:ppo-dril-hp}
\end{table}

\newpage

\section{Full Trajectory Optimization Algorithm}

\begin{algorithm}[h]
 \caption{\algo-SelectAction}
 \label{alg:full_trajopt}
\begin{algorithmic}[1]
 \small
\STATE{\textbf{SelectAction}}($s, \pi_{BC}, \hat{P}, r_{IL}, H, N, \sigma, k, K$):
    \STATE Set $\matr{R}^{N} = \vec{0}_N$ \Comment{This holds our N trajectory returns.}
    \STATE Set $\matr{A}_H^N = \vec{0}_H^N$ \Comment{This holds our N action trajectories of length H.}
    \FOR {$n=1..N$} 
        \STATE $l=n \mod K$ \Comment{Use consistent ensemble head throughout trajectory.}
	    \STATE $\hat{s}_1 = s$, $a_0 = T_0$, $R = 0$
        \FOR {$t=1..H$}
            \STATE $\epsilon \sim \mathcal{N}(0, \sigma^{2})$
            \STATE $\tilde{a}_t = \pi_{BC}^l(\hat{s}_t) + \epsilon$ \Comment{Sample current action using BC policy.}
            \STATE $\hat{s}_{t+1} = \hat{P}^l(\hat{s}_t, \sim{a}_t)$ \Comment{Sample next state from environment model.}
            \STATE $R^n = R^n + r_{IL}(\hat{s}_t, \sim{a}_t)$ \Comment{Take average reward over all ensemble members.}
            \STATE $\matr{A}_{t}^n = \tilde{a}_t$

        \ENDFOR
    \ENDFOR
    \STATE Sort $A_H^N$ according to $R^N$ descending.
    \STATE $a_1 =  1/k \sum_{n=1}^{k} A_1^n$ 
    \STATE{RETURN} $a_1$
\end{algorithmic}
\end{algorithm}

\newpage
\section{Effects of Initialization Noise in Environments}
\label{sec:init-noise-appendix}
\begin{figure}[ht]
     \centering
     \begin{subfigure}[b]{\textwidth}
         \centering
         \includegraphics[width=\textwidth]{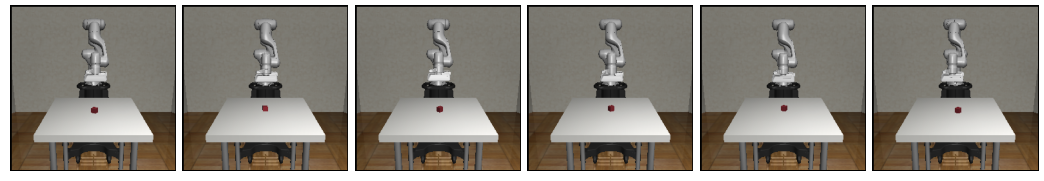}
     \end{subfigure}
     \hfill
     \begin{subfigure}[b]{\textwidth}
         \centering
         \includegraphics[width=\textwidth]{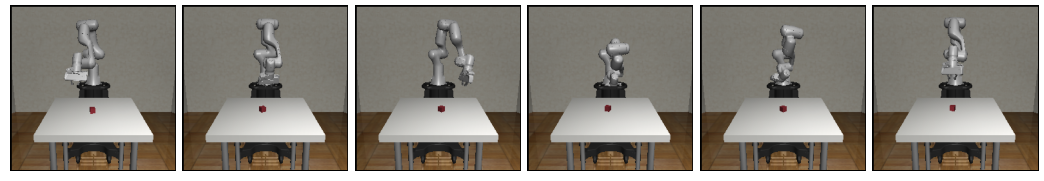}
         \caption{Lift: Default initialization noise of 0.02 on the top, and initialization noise of 0.4 on the bottom.}
         \label{fig:lift-noise}
     \end{subfigure}

     \centering
     \begin{subfigure}[b]{\textwidth}
         \centering
         \includegraphics[width=\textwidth]{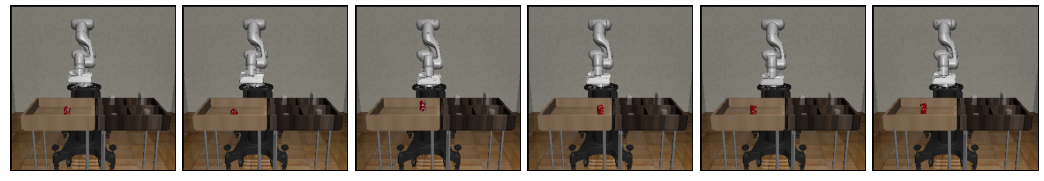}
     \end{subfigure}
     \hfill
     \begin{subfigure}[b]{\textwidth}
         \centering
         \includegraphics[width=\textwidth]{figures/fig_pick_0p4_init_noise.png}
         \caption{PickPlaceCan: Default initialization of 0.02 on the top, and initialization noise of 0.4 on the bottom.}
         \label{pick-noise}
     \end{subfigure}

     \centering
     \begin{subfigure}[b]{\textwidth}
         \centering
         \includegraphics[width=\textwidth]{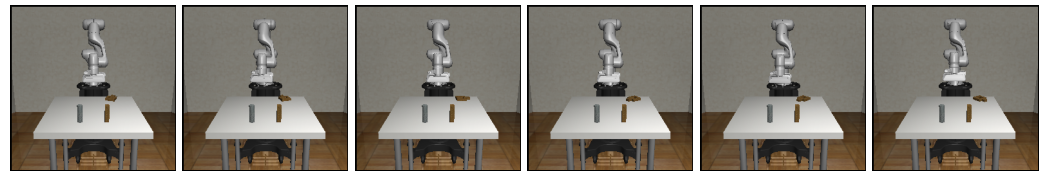}
     \end{subfigure}
     \hfill
     \begin{subfigure}[b]{\textwidth}
         \centering
         \includegraphics[width=\textwidth]{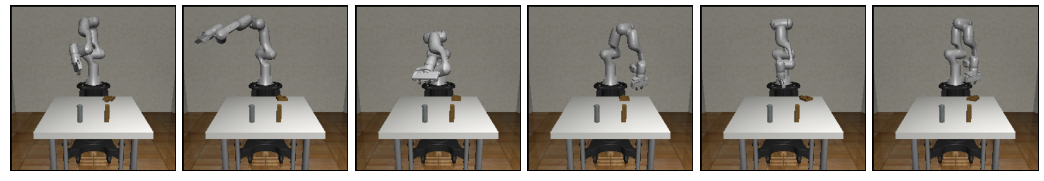}
         \caption{NutAssemblySquare: Default initialization noise of 0.02 on the top, and initialization noise of 0.4 on the bottom.}
         \label{nut-noise}
     \end{subfigure}

\caption{Randomly sampled initial states for Robosuite environments. For all figures, the top row contains initial states with the default initialization noise of 0.02, while the bottom row contains initial states with initialization noise of 0.4. Notice in Figure \ref{fig:lift-noise} that with higher initialization noise, the robotic arm is less likely to start in positions directly above the block.}
\label{fig:env_robosuite_init_noise_multi}
\end{figure}

We present the effects of initialization noise on the initial state of the robotic arm in Robomimic tasks. The noise value, $\sigma_{init}$, scales the standard deviation of Gaussian random noise applied to each of the robot's initial joint positions. The default $\sigma_{init}$ is 0.02, while we scaled up to 0.4 $\sigma_{init}$ in our experiments. The effect is demonstrated in Figure \ref{fig:env_robosuite_init_noise_multi}, where we sample 6 random initial states per environment for the default noise setting ($\sigma_{init} = 0.02$), and for the $\sigma_{init} = 0.4$ setting.

\newpage
\section{Emergent Retrying \& Videos}
\label{sec:emergent-retry}
\begin{figure}[h]
\centering
\includegraphics[width=\textwidth]{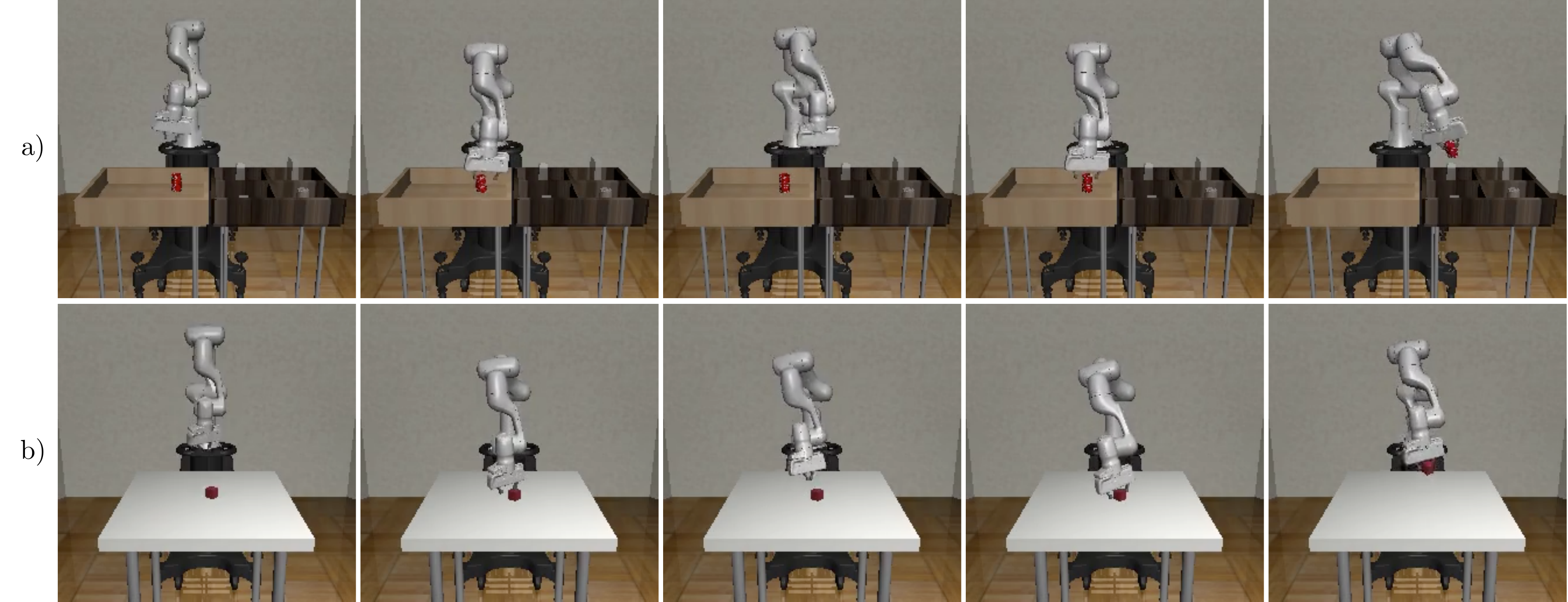}
\caption{Visualization of retries from the \algo agent in a) PickPlaceCan and b) Lift. We select a trajectory in each environment and show a subset of frames in a chronological order. Notice that the arm attempts to grab the object, moves upwards without the object, and comes back to successfully grab the object. BC does not showcase this behavior.}
\label{fig:retries}
\end{figure}

Figure~\ref{fig:retries} shows the retrying behavior of the POIR agent. We also attached multiple videos of the POIR-L2 agent (after 500k of finetuning) for the three environments. POIR-L2 is able to solve the task even when the initial position is very different from the one seen in the demonstrations. These videos also showcase the ability of POIR to retry the task until it is solved. In contrast, Ensemble BC is not able to retry the task and fail to solve it when the initial position is far from the ones in the demonstrations as shown in one of the attached videos. We also attached a video of DAC (after 5M steps) on Lift to showcase one failure mode of DAC: it keeps on re-trying the task without touching the object. 

\end{document}